\PassOptionsToPackage{table}{xcolor} 
\documentclass{article} 
\usepackage{iclr2027_conference,times}

\usepackage{amsmath,amsfonts,bm}

\def\eqref#1{equation~\ref{#1}}

\def\1{\bm{1}}

\DeclareMathAlphabet{\mathsfit}{\encodingdefault}{\sfdefault}{m}{sl}
\SetMathAlphabet{\mathsfit}{bold}{\encodingdefault}{\sfdefault}{bx}{n}

\usepackage{hyperref}
\usepackage{url}

\usepackage[utf8]{inputenc} 
\usepackage[T1]{fontenc}    
\usepackage{hyperref}       
\usepackage{url}            
\usepackage{booktabs}       
\usepackage{amsfonts}       
\usepackage{nicefrac}       
\usepackage{microtype}      
\usepackage[table]{xcolor}  
\usepackage{wrapfig}
\usepackage{placeins}
\usepackage{needspace}

\usepackage{algorithm}       
\usepackage{algorithmic}       

\usepackage{microtype}
\usepackage{graphicx}
\usepackage{subcaption}
\usepackage{booktabs} 
\usepackage{tikz}
\usepackage{pgfplots}
\pgfplotsset{compat=1.18}

\DeclareFixedFont{\ttb}{T1}{txtt}{bx}{n}{12} 
\DeclareFixedFont{\ttm}{T1}{txtt}{m}{n}{12}  

\usepackage{color}
\definecolor{deepblue}{rgb}{0,0,0.5}
\definecolor{deepred}{rgb}{0.6,0,0}
\definecolor{deepgreen}{rgb}{0,0.5,0}

\usepackage{listings}

\newcommand\pythonstyle{\lstset{
language=Python,
basicstyle=\small,
morekeywords={self},              
keywordstyle=\bf\color{deepblue},
emph={MyClass,__init__},          
emphstyle=\color{deepred},    
stringstyle=\color{deepgreen},
frame=tb,                         
showstringspaces=false,
}}

\lstnewenvironment{python}[1][]
{
\pythonstyle
\lstset{#1} 
}
{}

\usepackage{hyperref}

\usepackage{amsmath}
\usepackage{amssymb}
\usepackage{mathtools}
\usepackage{amsthm}

\usepackage{duckuments}

\usepackage[capitalize,noabbrev]{cleveref}

\theoremstyle{plain}

\theoremstyle{definition}

\theoremstyle{remark}

\usepackage[textsize=tiny]{todonotes}

  \title{The Token Games: Evaluating Language Model Reasoning with Puzzle Duels}

      \author{Simon Henniger\thanks{Equal contribution.}\\
      Harvard University\\
      \texttt{shenniger@fas.harvard.edu}
      \And
    Gabriel Poesia\footnotemark[1]\\
    University of Michigan, Ann Arbor\\
    \texttt{poesia@umich.edu}}

\newif\ifshowcomments
\showcommentstrue  

\iclrfinalcopy

\begin{document}

\maketitle

\begin{abstract}

Evaluating the reasoning capabilities of Large Language Models is increasingly challenging as models improve. Human curation of hard questions is highly expensive, especially in recent benchmarks using PhD-level domain knowledge to challenge the most capable models. Even then, there is always a concern about whether these questions test genuine reasoning or if similar problems have been seen during training. Here, we take inspiration from 16th-century mathematical duels to design \emph{The Token Games} (TTG): an evaluation framework where models challenge each other by creating their own puzzles. We leverage the format of Programming Puzzles --- given a function that returns a boolean, find inputs that make it return True --- to flexibly represent problems and enable verifying solutions. Using results from pairwise duels, we then compute Elo ratings, allowing us to compare models relative to each other. We evaluate 18 models --- frontier systems, low-reasoning-effort variants, and open-weights models --- on TTG, and closely match the rankings from existing benchmarks such as Humanity's Last Exam, without involving any human effort in creating puzzles and at a total cost of about \$2,400 --- a small fraction of the cost of sourcing expert-authored questions for comparable benchmarks. We also find that creating good puzzles is still a highly challenging task for current models. Overall, our work suggests new paradigms for evaluating reasoning that avoid saturation by design, and that allow testing models for other skills like creativity and task creation alongside problem solving.
\end{abstract}

\section{Introduction}

In the 16th century, mathematicians of the Italian renaissance rose to fame not by publishing breakthrough results but by winning public mathematical duels, in which two competitors posed a series of problems to each other: to win, one should solve the opponent's problems and stump them with one's own \citep{azzolini2019there}. In a famous duel against Antonio Maria Fior, Niccolò Tartaglia's victory revealed that he had found a general solution to cubic polynomials before any method was publicly known \citep{toscano2020secret}. As an evaluation of mathematical prowess, such duels had remarkable properties. Since solutions were verifiable, the public could confirm a victory without understanding the underlying reasoning. Since arbitrarily difficult problems were acceptable, as long as they were well-defined and the proposer could solve them, duels could not be saturated: there was always room for a new mathematician to showcase their brilliance. Finally, duels implicitly tested creativity: posing hard but already known problems would not work against opponents that had memorized the literature.

\begin{figure*}
    \centering
    \includegraphics[width=0.72\textwidth]{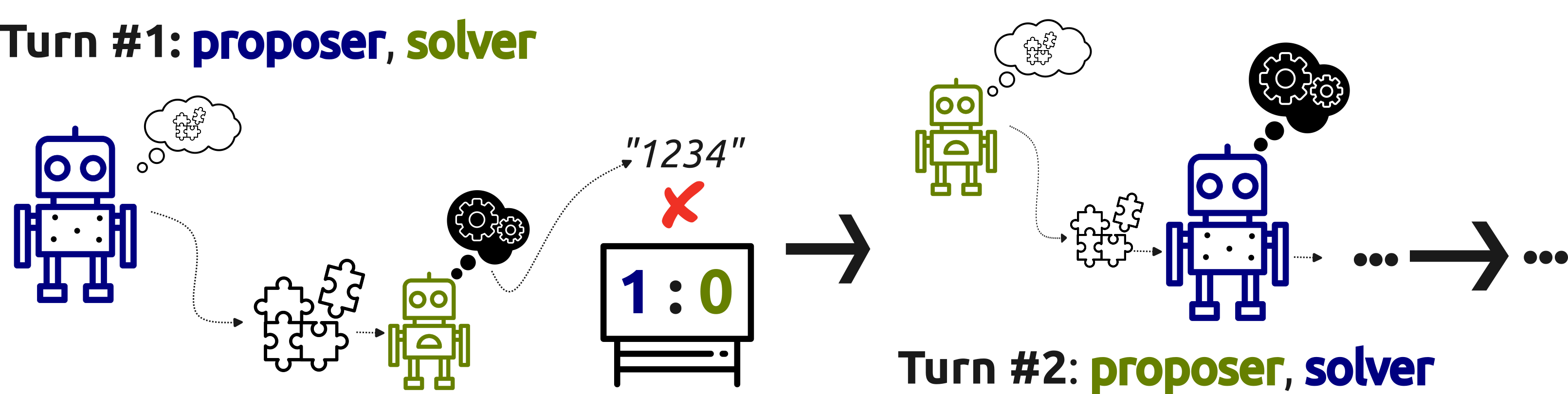}
    \caption{Reasoning duel in The Token Games. Two language models take turns between being puzzle proposers or solvers. Proposers score in a turn if they design a puzzle, give a correct solution, and have the opponent fail to solve the challenge. Puzzles are represented as Python functions returning a boolean value, with the challenge consisting of finding inputs to make it return true.}
    \label{fig:ttg}
\end{figure*}

In this paper, we take inspiration from mathematical duels to design a reasoning evaluation for Large Language Models that overcomes several limitations of current benchmarks. As LLMs improve in their capability for reasoning, it has become steadily harder for humans to design new questions that can measure progress as new models are released \citep{hle2024,rein2023gpqa}. We thus propose \emph{The Token Games} (TTG), a reasoning evaluation based on duel-style competitions between pairs of LLMs where models themselves propose challenge puzzles for each other. TTG inherits the properties outlined above: it cannot be saturated (stronger models can in principle always surpass previous ones), it evaluates models on arbitrarily difficult problems without expensive human curation, and it tests creativity: repeating hard problems seen during training is sub-optimal, as other models may have been trained on them too.

In a TTG duel, as illustrated in Figure~\ref{fig:ttg}, two language models take turns alternating between the roles of ``proposer'' and ``solver''. At the beginning of a turn, the proposer designs a puzzle and gives its own solution. To ensure that we can verify solutions, we adopt the format of Programming Puzzles \citep{schuster2021puzzles}: a puzzle is the source code of a verifier function that returns a boolean value, and a solution is an input for which it returns true. Despite its name, this representation is flexible enough to encode arbitrarily hard logical and mathematical problems without revealing their solutions (see Section~\ref{sec:ttg}). If the proposer's own solution is incorrect, the solver scores. If it is correct but the solver also succeeds, the turn ends in a draw; if the solver fails, the proposer scores. Whoever scores more points across a fixed number of turns wins the match.

We use the outcomes of a set of duels to compute Elo ratings for each language model, allowing us to rank models while always leaving space for new, more capable models to outperform. We empirically evaluate whether the TTG ranking agrees with other public benchmarks, such as GPQA-Diamond, HLE, and ARC-AGI. We find significant positive correlations between our Elo ratings and HLE ($\rho = 0.69$), ARC-AGI ($\rho=0.84$), GPQA Diamond ($\rho = 0.77$), SWE-Bench Pro ($\rho = 0.72$), and TextQuests (a text-adventure benchmark, $\rho=0.77$), all of which survive multiple-testing correction and a leave-one-out sensitivity analysis. Moreover, we also separately evaluate models' abilities in terms of the fraction of turns they won as a solver and as a proposer. We find that win rates as a proposer correlate most strongly with general reasoning benchmarks (up to $\rho = 0.90$ with ARC-AGI), and that win rates as a solver correlate most strongly with SWE-Bench Pro ($\rho = 0.80$). This indicates that, at a total cost of roughly \$2,400 for a full 18-model tournament (see Appendix~\ref{app:cost}) --- versus, e.g., the \$500k prize pool used to source the questions of HLE --- TTG is a strong proxy both for general reasoning (like HLE, ARC-AGI, GPQA) and code understanding (like SWE-Bench Pro). 

We show several analyses of the capabilities and failure modes that TTG reveals, including many cases of overconfidence of frontier models when designing hard puzzles (and failing to solve their own problems), as well as unique preferences for topics exhibited by different models across the 3,607 distinct puzzle topics arising in our tournament. Overall, our work contributes new ideas for designing scalable and inexpensive evaluations of LLM reasoning to stand the challenges posed by rapidly improving models.

\section{Related Work}

\paragraph{Language Model Reasoning.}
Early experiments with GPT-3 and PaLM \citep{wei2022chain} showed that prompting models to provide a \emph{chain-of-thought}, or fine-tuning them on scratchpads \citep{nye2021show}, substantially improves performance on multi-step reasoning problems. Recent models are increasingly trained with reinforcement learning targeted at reasoning in language \citep{zelikman2022star}, yielding a line of ``reasoning models'' starting with OpenAI o1 \citep{jaech2024openai} and DeepSeek r1 \citep{guo2025deepseek}. The resulting rapid, sometimes discontinuous jumps in benchmark performance have made measuring the capabilities of the most recent models a challenge of its own.

\paragraph{LLM Reasoning Benchmarks.}
Early benchmarks of mathematical problems used to evaluate LLMs included GSM8K \citep{cobbe2021training} and MATH \citep{hendrycks2021measuring}, extending to broader tasks such as BIG-Bench \citep{srivastava2023beyond} and ARC-Challenge \citep{clark2018think}. With rapid improvement, and sometimes near saturation, of performance in these benchmarks, recently benchmark developers turned to curating problems from experts, with benchmarks like GPQA \citep{rein2023gpqa}, Humanity's Last Exam (HLE, \citet{hle2024}), ARC-AGI \citep{arcagi} and Frontier Math \citep{glazer2024frontiermath} recruiting participant experts explicitly to design questions requiring highly specialized (e.g., PhD-level) knowledge in various fields. Relying on domain experts to design questions that are difficult for frontier models comes at a substantial cost: constructing HLE, for instance, involved sourcing experts with a prize pool of 500k USD for the top high-quality submissions. Even then, these benchmarks still risk being saturated by rapidly improving models: even one year after its release, for instance, the most performant models already achieved close to 90\% accuracy on GPQA. Furthermore, without having access to model's training data, it is impossible to disentangle factors such as potential data contamination (e.g., GPQA is available online) or similarity to previously existing problems (even if human authors were unaware of them). TextQuests \citep{textquests} and SWE-Bench Pro \citep{swebench}, which we also compare our rankings to, are both based on real-world data: old works of interactive fiction and source-code repositories from code hosting platforms, however, their difficulty is unlikely to scale well with newer models.

\paragraph{Problem Generation with LLMs.}
Several recent works use language models themselves to generate problems, but due to the concern that LLM-generated problems might be unreliable (e.g., have incorrect answers, missing hypotheses, or be ambiguous), these problems generally serve as training data rather than as a target for evaluation. Notable examples include MetaMath \citep{yu2024metamath}, which uses LLMs to generate synthetic mathematical problems for training in language, AbsoluteZero \citep{zhao2025absolute} which self-trained a Qwen model using model-generated programming puzzles, and models that generate and prove formal mathematical conjectures \citep{poesia2024learning,dong2025stp}. Our work uses LLMs to propose \emph{Programming Puzzles} \citep{schuster2021puzzles} for one another, in contrast with previous work where puzzles were extracted from human sources. Moreover, our main goal here is evaluation, whereas in all the aforementioned self-play setups models propose and train on their own puzzles.

\paragraph{LLM arenas and game-based evaluations}
Benchmarks such as Chatbot Arena \citep{chiang2024chatbot} and Copilot Arena \citep{chi2025copilot} also compare pairs of LLMs in competition, ranking their responses to user prompts. In existing ``arenas'', however, the judges are human participants, and the prompts are also created by users, unlike in our work where both generation and verification are automated. Other LLM evaluations place models in sequential decision-making setups, including in playing games such as in PuzzlePlex \citep{long2025puzzleplex} and SmartPlay \citep{wu2023smartplay}; here, however, puzzles are designed by benchmark authors, testing fixed capabilities, and might not scale in difficulty with increasingly capable models.

\section{The Token Games}
\label{sec:ttg}

We now describe \emph{The Token Games}, our proposed evaluation framework for LLM reasoning. 

TTG is based on the concept of a \emph{programming puzzle} introduced in \citet{schuster2021puzzles} as a unified representation for reasoning problems. A programming puzzle (or ``puzzle'', for short) is specified by a function $f(x)$ that receives arbitrary input $x$ and returns a boolean value (true or false). Then, $y$ is a solution to the puzzle if $f(y) = \texttt{true}$.

Specifying a verifier $f$ is a natural way to encode a wide variety of arbitrarily difficult reasoning challenges, including NP-complete problems such as boolean satisfiability and a range of mathematical problems; despite their name, programming puzzles can be extremely difficult while only requiring a very basic ability to read code (we quantify this in Section~\ref{sec:puzzle-tags}).
Even several open problems in mathematics, such as finding a cycle in the Collatz $3n + 1$ sequence, can be represented as a puzzle that simply \emph{verifies} the given solution, without $f$ revealing anything about how to solve the problem. Instead of collecting human-written puzzles to evaluate LLMs, our key idea is to design TTG around ``reasoning duels'' where two LLMs propose puzzles to challenge each other.

\begin{table*}[t]
\centering
\caption{Model performance comparison on The Token Games and, in column order, Humanity's Last Exam, ARC-AGI, SWE-Bench Pro, TextQuests, and GPQA Diamond. For TTG, we show match Elo with 95\% bootstrap confidence intervals as super/subscripts, the clean solver success rate on valid puzzles (Solv\%), the duel-scoring solver win rate (SolvW\%), and the proposer win rate (Prop\%), as defined in Section~\ref{sec:winrates}. For the other benchmarks, we report public-leaderboard accuracies (sources in Appendix~\ref{app:benchmark-results}). Each metric is followed by the model's rank; the top and bottom ranks are highlighted. The top-2 models are preserved (in order) in 97\% of bootstrap resamples, the same top-3 set in 61\%, and the same bottom-3 set in 73\%; bootstrap details in Appendix~\ref{app:additional}.}

\scriptsize
\setlength{\tabcolsep}{2pt}
\begin{tabular}{l|rr|rr|rr|rr|rr|rr|rr|rr|rr}
\toprule
& \multicolumn{8}{c|}{\textbf{TTG}} & \multicolumn{2}{c|}{\textbf{HLE}} & \multicolumn{2}{c|}{\textbf{ARC-AGI}} & \multicolumn{2}{c|}{\textbf{SWE-BP}} & \multicolumn{2}{c|}{\textbf{TQ}} & \multicolumn{2}{c}{\textbf{GPQA-D}} \\
\cmidrule(lr){2-9} \cmidrule(lr){10-11} \cmidrule(lr){12-13} \cmidrule(lr){14-15} \cmidrule(lr){16-17} \cmidrule(lr){18-19}
\textbf{Model} & \textbf{Elo} & \textbf{R} & \textbf{Solv\%} & \textbf{R} & \textbf{SolvW\%} & \textbf{R} & \textbf{Prop\%} & \textbf{R} & \textbf{Acc} & \textbf{R} & \textbf{Acc} & \textbf{R} & \textbf{Acc} & \textbf{R} & \textbf{Acc} & \textbf{R} & \textbf{Acc} & \textbf{R} \\
\midrule
gpt-\allowbreak 5.4 & 1746{\tiny$^{+213}_{-118}$} & \cellcolor{green!40}1 & 98.9 & \cellcolor{green!40}1 & 99.3 & \cellcolor{green!40}1 & 60.4 & \cellcolor{green!40}1 & 40.3 & 5 & 65.0 & 4 & 49.7 & 6 & 42.2 & \cellcolor{green!40}2 & 92.0 & 4 \\
gpt-\allowbreak 5.5 & 1445{\tiny$^{+120}_{-85}$} & \cellcolor{green!40}2 & 97.6 & \cellcolor{green!40}2 & 98.3 & \cellcolor{green!40}2 & 38.2 & \cellcolor{green!40}2 & 43.6 & \cellcolor{green!40}2 & 77.5 & \cellcolor{green!40}1 & 53.4 & 4 & 42.0 & \cellcolor{green!40}3 & 93.5 & \cellcolor{green!40}2 \\
claude-\allowbreak opus-\allowbreak 4-\allowbreak 8 & 1337{\tiny$^{+102}_{-85}$} & \cellcolor{green!40}3 & 88.2 & 7 & 92.3 & 6 & 28.7 & 4 & 42.2 & \cellcolor{green!40}3 & 67.5 & \cellcolor{green!40}3 & 65.4 & \cellcolor{green!40}1 & 40.1 & 4 & 92.0 & 4 \\
gemini-\allowbreak 3.1-\allowbreak pro-\allowbreak pre... & 1311{\tiny$^{+103}_{-87}$} & 4 & 85.7 & 8 & 90.1 & 8 & 33.8 & \cellcolor{green!40}3 & 45.9 & \cellcolor{green!40}1 & 73.3 & \cellcolor{green!40}2 & 46.7 & 7 & 45.8 & \cellcolor{green!40}1 & 94.1 & \cellcolor{green!40}1 \\
claude-\allowbreak opus-\allowbreak 4-\allowbreak 7 & 1280{\tiny$^{+102}_{-89}$} & 5 & 92.0 & \cellcolor{green!40}3 & 94.4 & \cellcolor{green!40}3 & 16.9 & 6 & 39.0 & 6 & 50.8 & 6 & 60.9 & \cellcolor{green!40}2 & 37.0 & 5 & 91.4 & 6 \\
claude-\allowbreak sonnet-\allowbreak 4-\allowbreak 6 & 1238{\tiny$^{+97}_{-84}$} & 6 & 88.5 & 6 & 91.6 & 7 & 13.3 & 8 & 21.1 & 11 & 24.2 & 8 & 53.8 & \cellcolor{green!40}3 & 31.5 & 8 & 87.5 & 11 \\
qwen/qwen3.7-\allowbreak max & 1203{\tiny$^{+86}_{-81}$} & 7 & 91.5 & 4 & 93.8 & 4 & 12.3 & 9 & -- & -- & -- & -- & -- & -- & -- & -- & 92.3 & \cellcolor{green!40}3 \\
gpt-\allowbreak 5.4-\allowbreak mini & 1096{\tiny$^{+101}_{-97}$} & 8 & 84.5 & 10 & 89.9 & 9 & 18.6 & 5 & 23.5 & 9 & 5.8 & 9 & 37.9 & 10 & 29.6 & 9 & 87.5 & 11 \\
claude-\allowbreak opus-\allowbreak 4-\allowbreak 7:low & 1045{\tiny$^{+79}_{-86}$} & 9 & 85.5 & 9 & 89.6 & 10 & 12.3 & 9 & -- & -- & -- & -- & -- & -- & -- & -- & -- & -- \\
grok-\allowbreak 4.20-\allowbreak 0309-\allowbreak rea... & 1045{\tiny$^{+90}_{-89}$} & 9 & 79.8 & 12 & 85.9 & 12 & 13.4 & 7 & 30.2 & 8 & 55.0 & 5 & 26.3 & \cellcolor{red!40}12 & 18.5 & \cellcolor{red!40}12 & 88.5 & 10 \\
claude-\allowbreak haiku-\allowbreak 4-\allowbreak 5 & 1000{\tiny$^{+0}_{-0}$} & 11 & 69.5 & 15 & 77.0 & 15 & 1.8 & \cellcolor{red!40}16 & 9.7 & \cellcolor{red!40}13 & 4.0 & \cellcolor{red!40}11 & 41.0 & 8 & 15.1 & \cellcolor{red!40}13 & 67.2 & \cellcolor{red!40}15 \\
z-\allowbreak ai/glm-\allowbreak 5.2 & 820{\tiny$^{+82}_{-113}$} & 12 & 80.8 & 11 & 86.0 & 11 & 8.1 & 13 & 41.4 & 4 & -- & -- & 51.8 & 5 & 33.3 & 7 & 89.5 & 9 \\
grok-\allowbreak 4-\allowbreak fast-\allowbreak reason... & 746{\tiny$^{+86}_{-132}$} & 13 & 62.0 & \cellcolor{red!40}16 & 71.0 & \cellcolor{red!40}16 & 6.0 & 15 & 17.8 & \cellcolor{red!40}12 & 3.3 & \cellcolor{red!40}12 & 12.0 & \cellcolor{red!40}13 & 20.1 & 11 & 84.7 & 13 \\
gpt-\allowbreak 5.5:low & 739{\tiny$^{+79}_{-118}$} & 14 & 89.3 & 5 & 92.6 & 5 & 9.5 & 12 & -- & -- & -- & -- & -- & -- & -- & -- & 91.0 & 7 \\
gemini-\allowbreak 3-\allowbreak flash-\allowbreak pre... & 725{\tiny$^{+97}_{-128}$} & 15 & 72.9 & 13 & 80.5 & 13 & 11.6 & 11 & 36.6 & 7 & 30.8 & 7 & 38.6 & 9 & 36.4 & 6 & 89.8 & 8 \\
deepseek-\allowbreak v3.2-\allowbreak thin... & 664{\tiny$^{+91}_{-147}$} & \cellcolor{red!40}16 & 71.9 & 14 & 79.7 & 14 & 6.9 & 14 & 21.8 & 10 & 5.0 & 10 & 33.1 & 11 & 21.2 & 10 & 84.0 & 14 \\
qwen/qwen3-\allowbreak coder & 210{\tiny$^{+31}_{-240}$} & \cellcolor{red!40}17 & 53.4 & \cellcolor{red!40}17 & 63.0 & \cellcolor{red!40}17 & 0.6 & \cellcolor{red!40}17 & -- & -- & -- & -- & -- & -- & -- & -- & 61.8 & \cellcolor{red!40}16 \\
gpt-\allowbreak 4o & 135{\tiny$^{+-3}_{-251}$} & \cellcolor{red!40}18 & 20.8 & \cellcolor{red!40}18 & 32.2 & \cellcolor{red!40}18 & 0.1 & \cellcolor{red!40}18 & 2.7 & \cellcolor{red!40}14 & 0.0 & \cellcolor{red!40}13 & 4.9 & \cellcolor{red!40}14 & 13.1 & \cellcolor{red!40}14 & 54.3 & \cellcolor{red!40}17 \\
\bottomrule
\end{tabular}
\setlength{\tabcolsep}{6pt}

\label{tab:rq1}
\vspace{-1.5em}
\end{table*}

\newpage
\subsection{Reasoning duel}


\begin{wrapfigure}{r}{0.5\linewidth}
\centering
\vspace{-1.5em}
\begin{minipage}{\linewidth}
\begin{algorithm}[H]
\caption{Reasoning Duel}
\begin{algorithmic}[1]
\label{algo:duel}

\REQUIRE Models $M_1, M_2$; number of rounds $R$

\STATE $S_1 \gets 0$, $S_2 \gets 0$, $H \gets \emptyset$
\FOR{$i \gets 1$ to $2R$}
    \STATE $m_{\mathrm{prop}} \gets M_{1 + (i \bmod 2)}$, $m_{\mathrm{solv}} \gets M_{1 + ((i+1) \bmod 2)}$
    \STATE $(f_i, s_i) \gets \textsc{ProposePuzzle}(m_{\mathrm{prop}}, H)$
    \STATE \textbf{if} $f_i(s_i) \neq \texttt{true}$: \textsc{ScorePoint}$(m_{\mathrm{solv}})$; \textbf{continue}
    \STATE $x_i \gets \textsc{Solve}(m_{\mathrm{solv}}, f_i)$
    \STATE \textbf{if} $f_i(x_i) = \texttt{false}$: \textsc{ScorePoint}$(m_{\mathrm{prop}})$
    \STATE $H \gets H \cup \{(f_i, \textsc{outcome}_i)\}$
\ENDFOR
\STATE \textbf{return} $M_1$ wins \textbf{if} $S_1 > S_2$; $M_2$ wins \textbf{if} $S_2 > S_1$; \textbf{else} draw
\end{algorithmic}
\end{algorithm}
\vspace{-1.5em}
\end{minipage}
\end{wrapfigure}

In TTG, pairs of LLMs in a pool compete in a series of \emph{duels}, which are conducted programatically as described in Algorithm~\ref{algo:duel}. In a duel, the two models play a fixed number of rounds where they alternate between the roles of \emph{proposer} and \emph{solver}. Each round can end with one of the two models scoring a point, or in a draw. Round $i$ starts with the proposer first designing a puzzle, $f_i$, and providing a solution, $s_i$, to its own puzzle. The proposer is given a private scratchpad for reasoning about their puzzle design before committing to a final puzzle and its answer; this allows models to engage in chain-of-thought reasoning when constructing puzzles. The puzzle is accepted only if the proposer's solution is correct: that is, $f_i(s_i) = \texttt{true}$. If this is not the case, the proposer immediately loses this round, and models change sides. Otherwise, if the proposer's solution is correct, the solver receives $f_i$ (but not the proposer's private reasoning) and is challenged to find some solution, $x_i$ (which need not be equal to $s_i$). We then test whether $f_i(x_i) = \texttt{true}$: if not, the proposer wins this round, and otherwise the round ends in a draw. At the end of all rounds, the model with most points is the winner. The match draws if both models end with the same number of points.

Solutions are verified by executing the puzzle code in a sandboxed Python environment. If evaluation of $f_i(s_i)$ or $f_i(x_i)$ results in a runtime error (e.g., due to malformed code, type errors, or timeout), the solution is treated as incorrect. A puzzle input can be any Python value that can be constructed in-place,
including strings, numbers and many data structures.

As the duel progresses, both models receive a history of all previous rounds. This allows models to adapt their strategies dynamically, such as adjusting to their opponent's difficulty level. Unlike for their own puzzles, each model only has access to partial information about their opponent's rounds: they see the puzzle code $f_i$, their own solution, and the round's outcome, but not the opponent's private reasoning or explanation of their puzzle's intended solution.

Reasoning duels have interesting properties from the perspective of LLM evaluations. Duels in principle incentivize models to creatively come up with \emph{novel questions}: an opponent might have already seen the solution of any puzzle taken straight from training data (indeed, we observe solvers defeating puzzles precisely by recognizing memorized artifacts; see Appendix~\ref{ex:genesis}). Second, the fact that models design the puzzles themselves allows them to show capabilities not explicitly asked for during benchmark design.
Finally, a duel requires a model to have a good self-evaluation of its own capabilities: it should not propose puzzles that it itself cannot solve. Our analysis in Section~\ref{sec:results} shows that even the most capable reasoning models are sometimes overconfident in this way.

\subsection{Rating models}
\label{sec:winrates}

To obtain a single scalar measure of reasoning ability across multiple models, we run duels between model pairs and fit ratings by applying the Bradley-Terry (BT) model on the set of outcomes, as in prior work on pairwise preference evaluation of LLMs \citep{chi2025copilot,chiang2024chatbot}. We parameterize the BT equation using the same exponential scoring function used in the Elo rating system, popular in chess and other competitive sports. Under the standard Elo model, the probability that model $A$ defeats $B$ in a match is $P(A \succ B) = 1/{\left(1 + 10^{(E_B - E_A)/\sigma}\right)}$,
where $E_A$ and $E_B$ denote the Elo ratings of models $A$ and $B$, respectively, and $\sigma$ is a scale parameter (we use $\sigma = 400$, the standard choice in chess).
We fit Elo ratings simultaneously on the final outcomes from all pairwise duels. Formally, let $w_{AB}$, $\ell_{AB}$, and $d_{AB}$ denote the number of wins, losses, and draws for model $A$ against model $B$ across all duels. Following the standard convention in Elo-based systems, we treat each draw as half a win for each player, defining effective wins $\tilde{w}_{AB} = w_{AB} + d_{AB}/2$ and effective losses $\tilde{\ell}_{AB} = \ell_{AB} + d_{AB}/2$. The negative log-likelihood of the observed outcomes is then
$\mathcal{L}(\{E_m\}) = -\sum_{(A,B)} \Big[ \tilde{w}_{AB} \log P(A \succ B) + \tilde{\ell}_{AB} \log P(B \succ A) \Big] \enspace.$

Since Elo ratings are only meaningful in relative terms, we fix the rating of an arbitrarily chosen model to $E_1 = 1000$ (using the model whose name comes first in lexicographical order)  and optimize the remaining ratings $\{E_2, \cdots, E_n\}$ to minimize $\mathcal{L}$ via gradient descent. A difference of $\sigma$ points corresponds to an expected win probability of approximately 91\% for the higher-rated model; each 100-point difference represents  a $\approx$ 64\% expected win rate.


Besides ranking models on their overall performance, in principle the ability of LLMs to solve hard puzzles need not be identical to their ability to propose good challenges. Thus, we also separately estimate each model's ability as a solver and as a proposer. To this end, we measure the fraction of \emph{rounds} won as a proposer (i.e., when playing as a proposer, how often did the model find a valid puzzle that its opponent could not solve). For the solver role we report two measures: the clean solver success rate (Solv\%), i.e., the fraction of \emph{valid} opposing puzzles the model solved, and the duel-scoring Solver Win Rate (SolvW\%), which counts all solver rounds and also credits the solver when the proposer forfeits the round with an invalid puzzle. 
We call the proposer-side measure the model's \emph{Proposer Win Rate} (Prop\%). Note that both solver measures consider the solver successful when it is able to solve the opponent's puzzle, whereas for computing the duel-level outcome we consider such situation a draw.

\section{Results}
\label{sec:results}

\begin{table}[b]
\centering
\caption{Spearman rank correlation ($\rho$) between model performance on TTG and reference benchmarks. Rows marked ``w/o incomplete rows'' recompute the correlations after excluding models that are missing scores on one or more of the benchmarks. Bolded results are for estimates with $p < 0.05$. Table~\ref{tab:robustness} in the appendix reports bootstrap confidence intervals, multiple-testing corrections, and a leave-one-out sensitivity analysis for these correlations.}
\footnotesize
\setlength{\tabcolsep}{3pt}
\begin{tabular}{l|cc|cc|cc|cc|cc}
\toprule
& \multicolumn{2}{c|}{\textbf{vs HLE}} & \multicolumn{2}{c|}{\textbf{vs ARC-AGI}} & \multicolumn{2}{c|}{\textbf{vs SWE-BP}} & \multicolumn{2}{c|}{\textbf{vs TQ}} & \multicolumn{2}{c}{\textbf{vs GPQA-D}} \\
\textbf{TTG Metric} & $\rho$ & $p$ & $\rho$ & $p$ & $\rho$ & $p$ & $\rho$ & $p$ & $\rho$ & $p$ \\
\midrule
Elo & \textbf{+0.69} & 0.007 & \textbf{+0.84} & 0.000 & \textbf{+0.72} & 0.004 & \textbf{+0.77} & 0.001 & \textbf{+0.77} & 0.000 \\
\quad (w/o incomplete rows) & \textbf{+0.78} & 0.002 & \textbf{+0.84} & 0.000 & \textbf{+0.78} & 0.002 & \textbf{+0.82} & 0.001 & \textbf{+0.83} & 0.000 \\
Solver Win Rate & \textbf{+0.70} & 0.005 & \textbf{+0.80} & 0.001 & \textbf{+0.80} & 0.001 & \textbf{+0.82} & 0.000 & \textbf{+0.83} & 0.000 \\
\quad (w/o incomplete rows) & \textbf{+0.75} & 0.003 & \textbf{+0.80} & 0.001 & \textbf{+0.81} & 0.001 & \textbf{+0.82} & 0.001 & \textbf{+0.79} & 0.001 \\
Proposer Win Rate & \textbf{+0.81} & 0.000 & \textbf{+0.90} & 0.000 & \textbf{+0.57} & 0.034 & \textbf{+0.85} & 0.000 & \textbf{+0.82} & 0.000 \\
\quad (w/o incomplete rows) & \textbf{+0.90} & 0.000 & \textbf{+0.90} & 0.000 & \textbf{+0.62} & 0.024 & \textbf{+0.88} & 0.000 & \textbf{+0.90} & 0.000 \\
\bottomrule
\end{tabular}

\label{tab:rq1-correlations}
\vspace{-2.5em}
\end{table}

We evaluated 18 models using TTG by running duels with all $18 \times 17 = 306$ ordered pairs of distinct models, repeating each ordered pairing three times to assess run-to-run stability (918 duels in total; see Table~\ref{tab:stability}). Each duel ran for 14 rounds, where models alternated roles (thus, each model played each role 7 times per duel and proposed 714 puzzles overall).
All duels can be found on URL \url{https://token-games.ai}. 
We compare TTG performance to performance on a number of benchmarks widely used to compare the strongest models: GPQA Diamond (\cite{rein2023gpqa}), TextQuests (\cite{textquests}), ARC-AGI v2 (\cite{arcagi}), SWE-Bench Pro (\cite{swebench}) and Humanity's Last Exam (HLE; \citet{hle2024}). We run models with the highest mode of ``thinking'' or ``reasoning'' enabled, except for the two marked low-effort variants (claude-opus-4-7:low and gpt-5.5:low) and gpt-4o as a non-reasoning baseline.
We investigate the following research questions:
\needspace{9\baselineskip}
\begin{description}
    \item[RQ1:]Does performance on TTG correlate with other standard reasoning benchmarks?
    \item[RQ2:]Do proposers show different preferences for the topics of their puzzles?
    \item[RQ3:]Are strong problem solvers also good problem proposers?
    \item[RQ4:]Do models use the duel's history to make puzzles more challenging in later rounds?
    \item[RQ5:] Do models fail to win as proposers for being too conservative (proposing easy puzzles) or too ambitious (failing on their own puzzle)?
    \item[RQ6:] How reproducible are TTG rankings across repeated tournaments?
\end{description}

\subsection{RQ1: Correlation between TTG and expert-authored benchmarks}

We first assess whether performance on TTG, a benchmarking framework with no human-designed problems, is indicative of strong performance on widely used reasoning benchmarks where questions were authored by diverse domain experts. Table~\ref{tab:rq1} shows three performance metrics based on TTG duels, as well as publicly reported accuracies on other benchmarks. Prop\%, Solv\%, and SolvW\% indicate the turn-level Proposer Win Rate, the clean solver success rate, and the duel-scoring Solver Win Rate respectively, as described in Section~\ref{sec:winrates}.

Despite the complete absence of human-authored problems in TTG, we observe that model rankings are generally consistent across benchmarks. The models leading TTG (GPT-5.4, GPT-5.5, Gemini 3.1 Pro Preview and Claude Opus 4-8) are also among the leaders on the expert-authored benchmarks, although TTG discriminates among them more sharply: GPT-5.4 and GPT-5.5 are nearly indistinguishable on HLE and GPQA-Diamond, yet separated by about 300 Elo points on TTG, with the top-2 ordering preserved in 97\% of bootstrap resamples (Table~\ref{tab:rq1}). More generally, the five TTG leaders span 29\% of the benchmark's full score range, versus 16\% on HLE and only 7\% on GPQA-Diamond: TTG retains discriminative power at the top of the range, where static benchmarks compress. At the other end of the spectrum, we find a similar effect: most models that struggle on TTG also tend to perform poorly on the expert-authored benchmarks. For instance, gpt-4o, our non-reasoning baseline, ranks in the bottom three on all rankings.

Table~\ref{tab:rq1-correlations} presents Spearman rank correlations between TTG metrics and the reference benchmarks. TTG Elo correlates positively and significantly with every benchmark ($\rho=0.69$ for HLE, $\rho=0.84$ for ARC-AGI, $\rho=0.72$ for SWE-Bench Pro, $\rho=0.77$ for TextQuests, and $\rho=0.77$ for GPQA-Diamond; all $p \le 0.007$), with higher values when restricting to models that have scores on every benchmark. All 15 correlations remain significant after Bonferroni correction and under leave-one-out deletion of any single model (Appendix Table~\ref{tab:robustness}); match Elo also ranks models near-identically to proposer Elo (Kendall $\tau=0.98$, differing only by two adjacent swaps). Proposer Win Rate correlates most strongly with the general reasoning benchmarks (up to $\rho=0.90$ for ARC-AGI) and most weakly with SWE-Bench Pro ($\rho=0.57$, $p=0.034$), whereas Solver Win Rate has its strongest relationship with SWE-Bench Pro ($\rho=0.80$). We find no same-lab advantage: a solver is no more likely to crack a puzzle proposed by a model from its own lab\footnote{Odds ratio $1.03$, $+0.5$ percentage points at the base rate; lab-relabeling permutation $p=0.87$), suggesting that duel outcomes are not driven by shared training pipelines or stylistic familiarity. We estimated this using a binomial logistic regression of round outcomes on proposer and solver fixed effects plus a same-lab indicator.}

In summary, we find a positive answer to our first research question: we find a \emph{statistically significant correlation between problem-solving performance in TTG, as measured by both solver and proposer win rates, and strong performance on expert-authored reasoning benchmarks}. This is striking due to the minimal cost in designing TTG, with no human-authored problems. Furthermore, we find that problem proposal and problem solving relate differently to existing evaluations, highlighting TTG's ability to separately assess these reasoning skills.

\subsection{RQ2: What are the puzzles about for different proposers?}
\label{sec:puzzle-tags}

To understand what TTG puzzles are about and whether different models exhibit diverse preferences of topics for their puzzles, we used \texttt{gpt-5.4-nano} with high reasoning to tag each puzzle in our annotated corpus with short topic tags (e.g., \texttt{modular-arithmetic}, \texttt{bit-manipulation}, \texttt{diophantine-equations}), given a list of 39 example tags, and reviewed the outputs for accuracy. Tags represent both broad (e.g., \texttt{number-theory}) and skill-specific categories (e.g., \texttt{chinese-remainder-theorem}), allowing us to understand topics at different levels of granularity, much like competitive programming problems are tagged in online platforms like Codeforces. This allowed us to obtain both a global understanding of puzzle topics, as well as a topic profile per proposer, which is summarized in Table~\ref{tab:proposer-tags}. On average, each puzzle had 5.58 tags associated with it --- puzzles frequently combine multiple topics by implementing a verifier that tests multiple, sometimes widely different, constraints.

\textbf{Top topics differ widely across models.} Although a few topics are globally common across proposers (the three most frequent tags globally are \texttt{modular-arithmetic} (49.0\% of puzzles), \texttt{strings} (44.6\%), and \texttt{bit-manipulation} (27.4\%)), the most frequent tag for each individual model still varies markedly: Gemini~3 Flash heavily uses \texttt{bit-manipulation} (80.7\% of its puzzles), GPT-5.5's most common tag is \texttt{cryptography} (50.7\%), GPT-5.4 stands out for \texttt{hashing} (48.3\%), and Claude Haiku and Claude Sonnet favor \texttt{number-theory} (54.6\% and 61.1\%).

\textbf{Each model has its own signature topic.} In Table~\ref{tab:proposer-tags}, the ``Distinctive preference'' column reports the tag from each proposer that had the highest ``lift'' (ratio between frequency in that proposer's puzzles and global frequency). Remarkably, these tags have \emph{no overlap} across the proposers, with lifts between $7.9\times$ and $19.0\times$ over the global rate. We see many striking examples of strong preferences not shared by other models: only Gemini~3.1 Pro writes \texttt{lambda-calculus} puzzles (11 in our corpus; no other proposer writes any; see Appendix~\ref{ex:church} for an example), and GPT-5.5 is responsible for \emph{all} 28 \texttt{elliptic-curve-cryptography}-tagged puzzles, an unusually large exclusive count consistent with its top tag (\texttt{cryptography}, 51\%). Other distinctive preferences include division-by-zero puzzles for Grok-4.20 (which proposed all six in the corpus), triangular-number puzzles for Claude Sonnet, RSA puzzles for DeepSeek, circular bit-rotations for Gemini~3 Flash, and systems of equations for GLM-5.2.

\textbf{Puzzle topics have a long tail.} Aside from the globally common topics, we also find a long tail of topics that different proposers contribute to TTG. Globally, most topics are rare: we find 2,918 tags (81\% of all tags; including date-time, involution, lexicographic-ordering) that only appear in 5 or fewer puzzles. The most frequent topic for each model that is only used by that proposer is reported on the last column of Table~\ref{tab:proposer-tags}. We find several niches that are filled by exactly one model, such as \texttt{slicing-permutation} from GPT-5.4, \texttt{utf-8-decoding} from DeepSeek, \texttt{two-complement} arithmetic from Qwen3-Coder, and \texttt{math-isqrt} from Claude Sonnet. Overall, we both observe (1) a wide range of topics arising from the simple structure of TTG, and (2) that proposers strongly show distinct preferences for particular topics besides common ones. Tags also track difficulty: against our three uniform solvers, \texttt{hmac} (36\%), \texttt{byte-manipulation} (38\%) and \texttt{cryptography} (45\%) puzzles have the lowest solve rates among common tags, while \texttt{digit-manipulation} (94\%) and \texttt{perfect-squares} (93\%) puzzles are the easiest --- consistent with hash- and cipher-style puzzles being the weapon of choice of the strongest proposers. We list diverse interesting examples in Appendix~\ref{app:examples}, and release our puzzle labels together with the data.

\textbf{Programming or reasoning?} A natural concern is that TTG, built on programming puzzles, measures programming skill while claiming to measure reasoning. However, no software-engineering topic appears among the most frequent tags. Instead, 85\% of puzzles carry at least one mathematics, cryptography, or self-reference tag. Puzzles encode diverse topics from Pell equations (Appendix~\ref{ex:sqtri}) to Church numerals (Appendix~\ref{ex:church}) to integer factorization (Appendix~\ref{ex:mersenne}). Second, the correlations of Table~\ref{tab:rq1-correlations} show that proposing correlates most strongly with general-reasoning benchmarks (ARC-AGI, $\rho=0.90$) and most weakly with the coding benchmark SWE-Bench Pro ($\rho=0.57$).

\subsection{RQ3: Are strong solvers also good proposers?}

Unlike existing evaluations, TTG allows us to measure both the aptitude for solving puzzles as well as for creating challenging ones. Here, we investigate how these abilities are related: do models that excel at solving puzzles also excel at proposing challenging problems?

We find a strong positive correlation between solver and proposer win rates (Spearman $\rho=0.81$, $p < 10^{-4}$). As shown in Table~\ref{tab:rq1}, models that rank highly as solvers tend to also rank highly as proposers: for instance, GPT-5.4 leads in both roles. This suggests that the reasoning capabilities required to solve difficult puzzles overlap with those needed to construct them.
However, we also observe a notable asymmetry when examining absolute performance levels. Across all rounds, proposers either fail to generate a valid puzzle, or to stump their opponents, 83.7\% of the time. Even the second-strongest proposer, GPT-5.5, wins only 38.2\% of rounds as a proposer, compared to a near-perfect 97.6\% as solver; GPT-5.4, the strongest proposer, is the sole outlier at 60.4\%. At the other extreme, Claude Haiku 4.5 wins only 1.8\% of its rounds as proposer, despite itself solving 69.5\% of the valid puzzles proposed by its opponents (see Table~\ref{combined}). The stumping puzzles are nonetheless genuinely solvable: 95.0\% of all valid puzzles were solved by \emph{someone} --- either by the opponent during the duel, or by the proposer itself when re-solving its puzzle cold, without access to its private notes (``Either S'' in Table~\ref{combined}) --- and the remaining 5\% concentrate among the strongest proposers' hardest puzzles, consistent with genuine difficulty rather than unsolvability.

This disparity reveals that \textbf{creating good puzzles is substantially harder than solving them}. Models frequently either propose puzzles that their opponents can solve, or fail to solve their own problems. The proposer role thus also represents a highly discriminating test of reasoning capability, with a wider spread in performance across models.

\subsection{RQ4: Can proposers increase difficulty using the duel's history?}

In TTG duels, proposers can see the whole history of interactions. This design decision was based on the hypothesis that proposers should adapt the difficulty of their problems as they observe either their own failures in proposing valid puzzles with correct solutions (likely indicating they should be less ambitious) or seeing the opponent's successes as solvers (pressuring them to attempt to propose harder puzzles). To assess whether such an adaptation is taking place, we use the performance of three uniform solvers --- Qwen3.8 27B (run locally), Grok 4 Fast Reasoning, and GPT-5.4 Nano, spanning overall solve rates from 57\% to 81\% --- on all puzzles extracted from all duels as an assessment of puzzle difficulty beyond the two competing models. When solving a puzzle, as also happens during duels, models are only given general instructions on format and the puzzle itself, but no other previous context. We evaluate each solver's success rate for puzzles proposed across turns from 1 to 14 in all duels.

Figure~\ref{fig:solve-by-turn-comparison} shows the results. Solve rates decline across turns for all three solvers, and the sensitivity is monotone in solver strength: the weakest solver (Grok 4 Fast) loses 0.81 percentage points of solve rate per turn --- over 10 points across a duel --- while the strongest (GPT-5.4 Nano) loses only 0.33 points per turn. The duels themselves show the same trend: the number of puzzles solved by the duel opponent falls from 594 in the first turn to 490 in the last (Appendix Table~\ref{byturn}). Overall, we find that proposers measurably increase difficulty across turns, and this is felt most strongly by weaker solvers. The uniform solvers also validate the duel outcomes themselves: their per-proposer failure rates agree closely with each other and with the rate at which each proposer stumped its opponents in the arena (Spearman $\rho$ between $0.87$ and $0.98$ per solver, and $\rho=0.99$ for their average; $n=18$ proposers), indicating that duel outcomes measure puzzle difficulty rather than opponent-specific quirks.

\subsection{RQ5: Are failing proposers too conservative or too ambitious?}

When playing as a proposer, a model can fail to win the turn in one of two ways: (1) its own solution is wrong (giving a point to its opponent), or (2) the opponent's solution is correct (ending the turn in a draw). In the former case, the proposer was arguably overly confident in its ability to solve the puzzle it proposed. In the latter case, the puzzle was too easy for the opponent. How do models behave with respect to these two failure modes as proposers?

\begin{figure}[t]
  \centering
  \begin{minipage}{0.51\linewidth}
    \centering
    \vspace{0.5em}
    \includegraphics[width=\linewidth]{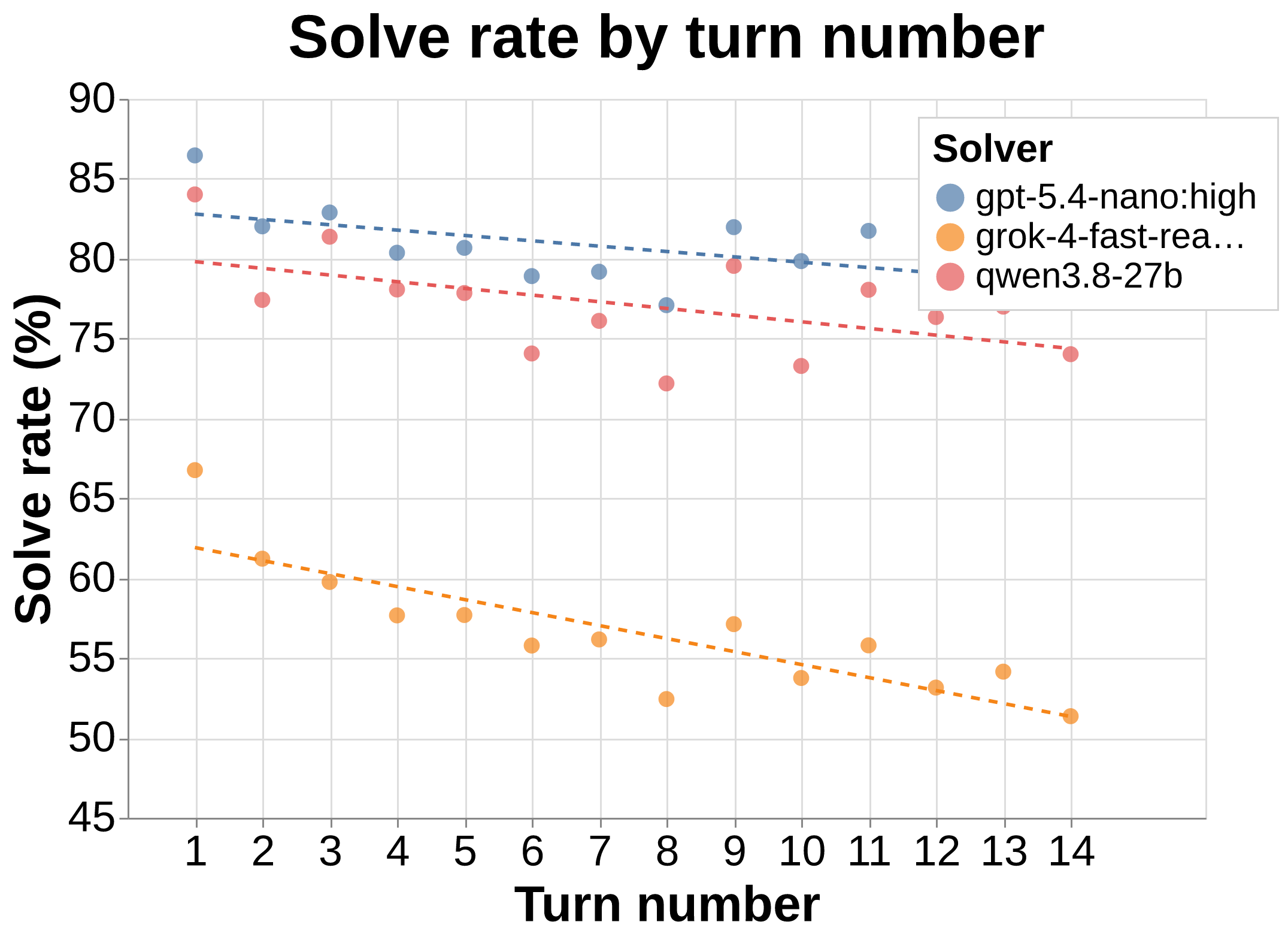}
    \vspace{0.5em}
    \caption{Solve rates of three uniform solvers (Qwen3.8 27B, Grok 4 Fast Reasoning, and GPT-5.4 Nano) on puzzles from all duels, by the turn in which the puzzle was proposed. Regression fits show puzzles becoming slightly harder each turn.}
    \label{fig:solve-by-turn-comparison}
  \end{minipage}
  \hfill
  \begin{minipage}{0.48\linewidth}
    \centering
    \includegraphics[width=\linewidth]{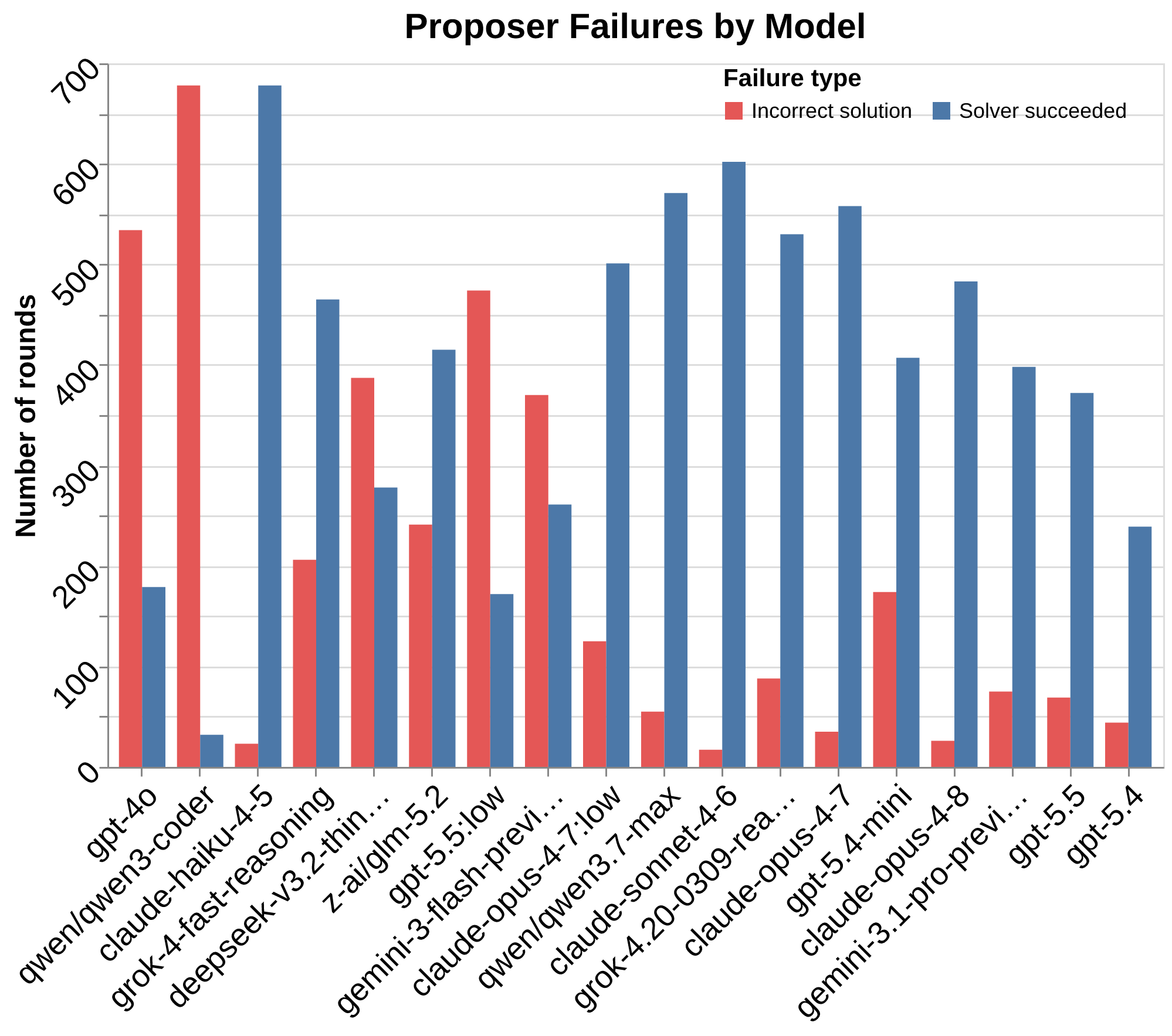}
    \caption{Failure modes for proposers. Did the model's own sample solution fail (red) or did the solver succeed in solving the puzzle (blue)? Each model had 714 puzzle proposals.}
    \label{fig:proposer-failures}
  \end{minipage}
  \vspace{-1em}
\end{figure}

Figure~\ref{fig:proposer-failures} shows the number of failures as proposers of each of these two categories and Table~\ref{combined} shows overall performance as a proposer. Across all models, we see that failing at providing hard-enough puzzles is more common (55.6\% of all puzzle proposals are solved by the opponent) than providing a too-hard puzzle with an incorrect sample solution (28.2\%). Data, however, differs a lot by model. Gemini 3 Flash Preview exhibits high overconfidence (51.2\% penalties for an incorrect solution to its own puzzle).

Sonnet 4.6, on the other hand, usually writes puzzles that are too easy -- its competitors can solve Sonnet's puzzles 82.0\% of the time. We find recurring failure patterns: arithmetic slips in the proposer's own construction (Appendix~\ref{ex:pell61}, \ref{ex:minirsa}), mismatches between the proposer's mathematical model and Python's actual semantics, and subtle representation issues such as string re-escaping (Appendix~\ref{ex:quine49}).


\subsection{RQ6: How reproducible are TTG rankings?}

Repeating every ordered pairing three times lets us measure reproducibility directly. At the match level, 61\% of pairings produce the identical outcome in all three repetitions and only 17.6\% ever reverse; the mean within-pairing variance of the match score is 0.049, against 0.25 for a coin flip (Appendix Table~\ref{tab:stability}). At the ranking level, the bootstrap intervals of Table~\ref{tab:rq1} preserve the top-2 ordering in 97\% of resamples, with uncertainty concentrated mid-table. At the tournament level, the three repetitions form independent tournaments whose Elo rankings agree at Spearman $\rho=0.95$ pairwise, implying a Spearman--Brown reliability of 0.98 for the full tournament. TTG's own noise therefore attenuates the correlations of Table~\ref{tab:rq1-correlations} by at most 0.01: the gap to static benchmarks reflects differences in what is measured, not ranking instability --- and a single repetition already suffices to recover the ranking, at a third of our reported cost.

\section{Conclusion}
We propose a competition-style evaluation of LLM reasoning with no human-authored problems. For 18 models, performance on TTG correlates significantly with other standard reasoning benchmarks that were highly expensive to design.
With 18 models, we were able to run duels between all pairs of models (three repetitions of each ordered pairing). In the future, BT ratings can be inferred from sparse subsets of all duels. This will become important as we add new models to TTG and dueling against all previous models becomes infeasible.

\FloatBarrier
\subsection*{AI Disclosure Statement}
We used generative AI tools for assisting with code implementation, data analysis/visualization, proof-reading and drafting of the paper. LLMs are also the object of our analysis. Generative AI tools were not used for brainstorming, suggesting experiments or hypotheses, or other higher-level research tasks. We have reviewed all AI-assisted work. We take full responsibility for the final content of this work, including text and artifacts produced with the aid of generative AI.

\section*{Acknowledgments}
We thank Trevor DePodesta for his contributions. This work has been made possible in part by a Berkman Klein Center Ideas-to-Impact Incubator grant, a gift from the Chan Zuckerberg Initiative Foundation to establish the Kempner Institute for the Study of Natural and Artificial Intelligence at Harvard University and by a grant from the Harvard Mind Brain Behavior Interfaculty Initiative. Simon Henniger was supported by the Harvard John A. Paulson School of Engineering and Applied Sciences (SEAS) Prize Fellowship and the German Academic Fellowship Organization, funded by the German Federal Ministry for Economic Affairs and Energy.

\bibliography{references}

@inproceedings{wei2022chain,
  title={Chain-of-Thought Prompting Elicits Reasoning in Large Language Models},
  author={Wei, Jason and Wang, Xuezhi and Schuurmans, Dale and Bosma, Maarten and Ichter, Brian and Xia, Fei and Chi, Ed H. and Le, Quoc V. and Zhou, Denny},
  booktitle={Advances in Neural Information Processing Systems},
  volume={35},
  year={2022}
}

@article{nye2021show,
  title={Show Your Work: Scratchpads for Intermediate Computation with Language Models},
  author={Nye, Maxwell and Andreassen, Anders Johan and Gur-Ari, Guy and Michalewski, Henryk and Austin, Jacob and Bieber, David and Dohan, David and Lewkowycz, Aitor and Bosma, Maarten and Luan, David and Sutton, Charles and Odena, Augustus},
  journal={arXiv preprint arXiv:2112.00114},
  year={2021}
}

@article{cobbe2021training,
  title={Training Verifiers to Solve Math Word Problems},
  author={Cobbe, Karl and Kosaraju, Vineet and Bavarian, Mohammad and Chen, Mark and Jun, Heewoo and Kaiser, Lukasz and Plappert, Matthias and Tworek, Jerry and Hilton, Jacob and Nakano, Reiichiro and Hesse, Christopher and Schulman, John},
  journal={arXiv preprint arXiv:2110.14168},
  year={2021}
}

@inproceedings{hendrycks2021measuring,
  title={Measuring Mathematical Problem Solving With the {MATH} Dataset},
  author={Hendrycks, Dan and Burns, Collin and Kadavath, Saurav and Arora, Akul and Basart, Steven and Tang, Eric and Song, Dawn and Steinhardt, Jacob},
  booktitle={Thirty-fifth Conference on Neural Information Processing Systems Datasets and Benchmarks Track},
  year={2021}
}

@article{srivastava2023beyond,
  title={Beyond the Imitation Game: Quantifying and extrapolating the capabilities of language models},
  author={{BIG-bench authors}},
  journal={Transactions on Machine Learning Research},
  issn={2835-8856},
  year={2023},
  url={https://openreview.net/forum?id=uyTL5Bvosj}
}

@article{clark2018think,
  title={Think you have Solved Question Answering? Try {ARC}, the {AI2} Reasoning Challenge},
  author={Clark, Peter and Cowhey, Isaac and Etzioni, Oren and Khot, Tushar and Sabharwal, Ashish and Schoenick, Carissa and Tafjord, Oyvind},
  journal={arXiv preprint arXiv:1803.05457},
  year={2018}
}

@article{rein2023gpqa,
  title={{GPQA}: A Graduate-Level Google-Proof Q\&A Benchmark},
  author={Rein, David and Hou, Betty Li and Stickland, Asa Cooper and Petty, Jackson and Pang, Richard Yuanzhe and Dirani, Julien and Michael, Julian and Bowman, Samuel R.},
  journal={arXiv preprint arXiv:2311.12022},
  year={2023}
}

@article{hle2024,
  title={Humanity's Last Exam},
  author={Phan, Long and Gatti, Alice and Han, Ziwen and Li, Nathaniel and Hu, Josephina and Zhang, Hugh and {Scale AI} and {Center for AI Safety}},
  journal={arXiv preprint arXiv:2501.14249},
  year={2025}
}

@inproceedings{yu2024metamath,
  title={{MetaMath}: Bootstrap Your Own Mathematical Questions for Large Language Models},
  author={Yu, Longhui and Jiang, Weisen and Shi, Han and Yu, Jincheng and Liu, Zhengying and Zhang, Yu and Kwok, James T. and Li, Zhenguo and Weller, Adrian and Liu, Weiyang},
  booktitle={International Conference on Learning Representations},
  year={2024}
}

@inproceedings{zelikman2022star,
  title={{STaR}: Bootstrapping Reasoning With Reasoning},
  author={Zelikman, Eric and Wu, Yuhuai and Mu, Jesse and Goodman, Noah},
  booktitle={Advances in Neural Information Processing Systems},
  volume={35},
  year={2022}
}

@inproceedings{schuster2021puzzles,
  title={Programming Puzzles},
  author={Schuster, Tal and Kalyan, Ashwin and Polozov, Oleksandr and Kalai, Adam Tauman},
  booktitle={Thirty-fifth Conference on Neural Information Processing Systems Datasets and Benchmarks Track},
  year={2021}
}

@article{azzolini2019there,
  title={There Were No Medals To Be Won: Scientific Duels in the Italian Renaissance},
  author={Azzolini, Monica},
  journal={Nuncius},
  volume={34},
  number={2},
  pages={258--283},
  year={2019},
  publisher={Brill}
}

@book{toscano2020secret,
  title={The Secret Formula: How a Mathematical Duel Inflamed Renaissance Italy and Uncovered the Cubic Equation},
  author={Toscano, Fabio},
  year={2020},
  publisher={Princeton University Press}
}

@article{jaech2024openai,
  title={Openai o1 system card},
  author={Jaech, Aaron and Kalai, Adam and Lerer, Adam and Richardson, Adam and El-Kishky, Ahmed and Low, Aiden and Helyar, Alec and Madry, Aleksander and Beutel, Alex and Carney, Alex and others},
  journal={arXiv preprint arXiv:2412.16720},
  year={2024}
}

@article{guo2025deepseek,
  title={DeepSeek-R1 incentivizes reasoning in LLMs through reinforcement learning},
  author={Guo, Daya and Yang, Dejian and Zhang, Haowei and Song, Junxiao and Wang, Peiyi and Zhu, Qihao and Xu, Runxin and Zhang, Ruoyu and Ma, Shirong and Bi, Xiao and others},
  journal={Nature},
  volume={645},
  number={8081},
  pages={633--638},
  year={2025},
  publisher={Nature Publishing Group UK London}
}

@article{glazer2024frontiermath,
  title={Frontiermath: A benchmark for evaluating advanced mathematical reasoning in ai},
  author={Glazer, Elliot and Erdil, Ege and Besiroglu, Tamay and Chicharro, Diego and Chen, Evan and Gunning, Alex and Olsson, Caroline Falkman and Denain, Jean-Stanislas and Ho, Anson and Santos, Emily de Oliveira and others},
  journal={arXiv preprint arXiv:2411.04872},
  year={2024}
}

@article{dong2025stp,
  title={Stp: Self-play llm theorem provers with iterative conjecturing and proving},
  author={Dong, Kefan and Ma, Tengyu},
  journal={arXiv preprint arXiv:2502.00212},
  year={2025}
}

@article{poesia2024learning,
  title={Learning formal mathematics from intrinsic motivation},
  author={Poesia, Gabriel and Broman, David and Haber, Nick and Goodman, Noah},
  journal={Advances in Neural Information Processing Systems},
  volume={37},
  pages={43032--43057},
  year={2024}
}

@inproceedings{chiang2024chatbot,
  title={Chatbot arena: An open platform for evaluating llms by human preference},
  author={Chiang, Wei-Lin and Zheng, Lianmin and Sheng, Ying and Angelopoulos, Anastasios Nikolas and Li, Tianle and Li, Dacheng and Zhu, Banghua and Zhang, Hao and Jordan, Michael and Gonzalez, Joseph E and others},
  booktitle={Forty-first International Conference on Machine Learning},
  year={2024}
}

@article{chi2025copilot,
  title={Copilot arena: A platform for code llm evaluation in the wild},
  author={Chi, Wayne and Chen, Valerie and Angelopoulos, Anastasios Nikolas and Chiang, Wei-Lin and Mittal, Aditya and Jain, Naman and Zhang, Tianjun and Stoica, Ion and Donahue, Chris and Talwalkar, Ameet},
  journal={arXiv preprint arXiv:2502.09328},
  year={2025}
}

@inproceedings{
swebench,
title={{SWE}-bench: Can Language Models Resolve Real-world Github Issues?},
author={Carlos E Jimenez and John Yang and Alexander Wettig and Shunyu Yao and Kexin Pei and Ofir Press and Karthik R Narasimhan},
booktitle={The Twelfth International Conference on Learning Representations},
year={2024},
url={https://openreview.net/forum?id=VTF8yNQM66}
}

@misc{textquests,
      title={TextQuests: How Good are LLMs at Text-Based Video Games?}, 
      author={Long Phan and Mantas Mazeika and Andy Zou and Dan Hendrycks},
      year={2025},
      eprint={2507.23701},
      archivePrefix={arXiv},
      primaryClass={cs.AI},
      url={https://arxiv.org/abs/2507.23701}, 
}

@misc{arcagi,
      title={ARC-AGI-2: A New Challenge for Frontier AI Reasoning Systems}, 
      author={Francois Chollet and Mike Knoop and Gregory Kamradt and Bryan Landers and Henry Pinkard},
      year={2026},
      eprint={2505.11831},
      archivePrefix={arXiv},
      primaryClass={cs.AI},
      url={https://arxiv.org/abs/2505.11831}, 
}

@article{zhao2025absolute,
  title={Absolute zero: Reinforced self-play reasoning with zero data},
  author={Zhao, Andrew and Wu, Yiran and Yue, Yang and Wu, Tong and Xu, Quentin and Lin, Matthieu and Wang, Shenzhi and Wu, Qingyun and Zheng, Zilong and Huang, Gao},
  journal={arXiv preprint arXiv:2505.03335},
  year={2025}
}

@article{long2025puzzleplex,
  title={PuzzlePlex: Benchmarking Foundation Models on Reasoning and Planning with Puzzles},
  author={Long, Yitao and Jiang, Yuru and Liu, Hongjun and Zhao, Yilun and Sun, Jingchen and Shen, Yiqiu and Zhao, Chen and Cohan, Arman and Shasha, Dennis},
  journal={arXiv preprint arXiv:2510.06475},
  year={2025}
}

@article{wu2023smartplay,
  title={Smartplay: A benchmark for llms as intelligent agents},
  author={Wu, Yue and Tang, Xuan and Mitchell, Tom M and Li, Yuanzhi},
  journal={arXiv preprint arXiv:2310.01557},
  year={2023}
}
\bibliographystyle{iclr2027_conference}

\newpage
\appendix
\onecolumn
\section{Additional Results}
\label{app:additional}

The Elo confidence intervals in Table~\ref{tab:rq1} are computed from a 1000-sample bootstrap over match outcomes; intervals are relative to the anchor model (claude-haiku-4-5, fixed at 1000), whose own interval is degenerate by construction.

Table~\ref{tab:proposer-tags} shows the per-proposer puzzle tag profile discussed in Section~\ref{sec:puzzle-tags}.

\begin{table}[htbp]
\centering
\caption{Per-proposer puzzle tag profile across all duels. \emph{Distinctive preference}: among tags appearing $\geq 5$ times for the proposer, the one with the largest ``lift'', i.e., the ratio of the tag's per-proposer frequency to its global frequency across all proposers; an $N\times$ multiplier means the proposer uses the tag $N$ times more often than the average proposer. \emph{Exclusive tag}: the most-used tag whose appearances \emph{all} come from that one proposer (puzzle count in parentheses).}
\label{tab:proposer-tags}
\scriptsize
\setlength{\tabcolsep}{4pt}
\begin{tabular}{@{}l p{4.9cm} p{3.6cm}@{}}
\toprule
\textbf{Proposer} & \textbf{Distinctive preference} & \textbf{Exclusive tag} \\
\midrule
claude-\allowbreak{}haiku-\allowbreak{}4-\allowbreak{}5 & \texttt{factorials} (1.4\%, 12.9$\times$) & \texttt{arithmetic-\allowbreak{}exponentiation} (3) \\
claude-\allowbreak{}opus-\allowbreak{}4-\allowbreak{}7 & \texttt{string-\allowbreak{}to-\allowbreak{}digits} (2.1\%, 5.9$\times$) & \texttt{involution} (2) \\
claude-\allowbreak{}opus-\allowbreak{}4-\allowbreak{}7:low & \texttt{range-\allowbreak{}checking} (0.7\%, 9.0$\times$) & \texttt{arithmetic-\allowbreak{}algebra} (2) \\
claude-\allowbreak{}opus-\allowbreak{}4-\allowbreak{}8 & \texttt{modular-\allowbreak{}square-\allowbreak{}roots} (0.7\%, 12.9$\times$) & \texttt{code-\allowbreak{}points} (2) \\
claude-\allowbreak{}sonnet-\allowbreak{}4-\allowbreak{}6 & \texttt{prime-\allowbreak{}testing} (1.3\%, 8.1$\times$) & \texttt{math-\allowbreak{}isqrt} (3) \\
deepseek-\allowbreak{}v3.2-\allowbreak{}thinking & \texttt{rsa} (1.3\%, 7.7$\times$) & \texttt{base64-\allowbreak{}encoding} (2) \\
gemini-\allowbreak{}3-\allowbreak{}flash-\allowbreak{}preview & \texttt{circular-\allowbreak{}rotation} (2.8\%, 16.4$\times$) & \texttt{unsigned-\allowbreak{}integer} (3) \\
gemini-\allowbreak{}3.1-\allowbreak{}pro-\allowbreak{}preview & \texttt{lambda-\allowbreak{}calculus} (1.5\%, 18.0$\times$) & \texttt{lambda-\allowbreak{}calculus} (11) \\
gpt-\allowbreak{}4o & \texttt{case-\allowbreak{}insensitive-\allowbreak{}comparison} (0.7\%, 12.9$\times$) & \texttt{vowel-\allowbreak{}detection} (2) \\
gpt-\allowbreak{}5.4 & \texttt{slicing-\allowbreak{}permutation} (0.7\%, 18.0$\times$) & \texttt{slicing-\allowbreak{}permutation} (5) \\
gpt-\allowbreak{}5.4-\allowbreak{}mini & \texttt{cipher} (1.5\%, 11.6$\times$) & \texttt{list-\allowbreak{}indexing} (4) \\
gpt-\allowbreak{}5.5 & \texttt{elliptic-\allowbreak{}curve-\allowbreak{}cryptography} (3.8\%, 18.0$\times$) & \texttt{elliptic-\allowbreak{}curve-\allowbreak{}cryptography} (27) \\
gpt-\allowbreak{}5.5:low & \texttt{hash-\allowbreak{}like-\allowbreak{}mixing} (0.8\%, 10.8$\times$) & \texttt{obfuscated-\allowbreak{}validation} (3) \\
grok-\allowbreak{}4-\allowbreak{}fast-\allowbreak{}reasoning & \texttt{ord-\allowbreak{}chr} (1.4\%, 8.2$\times$) & \texttt{polynomial-\allowbreak{}iteration} (3) \\
grok-\allowbreak{}4.20-\allowbreak{}0309-\allowbreak{}reasoning & \texttt{division-\allowbreak{}by-\allowbreak{}zero} (0.7\%, 18.0$\times$) & \texttt{division-\allowbreak{}by-\allowbreak{}zero} (5) \\
qwen/qwen3-\allowbreak{}coder & \texttt{binary-\allowbreak{}parsing} (1.8\%, 13.0$\times$) & \texttt{two-\allowbreak{}complement} (3) \\
qwen/qwen3.7-\allowbreak{}max & \texttt{ascii-\allowbreak{}ord-\allowbreak{}mapping} (1.0\%, 14.0$\times$) & \texttt{leap-\allowbreak{}year-\allowbreak{}calculation} (2) \\
z-\allowbreak{}ai/glm-\allowbreak{}5.2 & \texttt{systems-\allowbreak{}of-\allowbreak{}equations} (1.1\%, 7.6$\times$) & \texttt{equation-\allowbreak{}system} (2) \\
\bottomrule
\end{tabular}

\end{table}

Table~\ref{solve-cmp} shows puzzle difficulty by each model as a proposer, as evaluated by the ability of three uniform solvers (Qwen3.8 27B, Grok 4 Fast Reasoning, and GPT-5.4 Nano) to solve them.

\begin{table*}[htbp]
        \caption{Uniform solvers by proposer model. After completing all duels, we mined puzzles from their logs and had three models attempt them outside the duel context: Qwen3.8 27B, Grok 4 Fast Reasoning, and GPT-5.4 Nano. We only used valid puzzles, i.e. those that had correct sample solutions, and we show each solver's solve rate for puzzles by their original proposer.}
  \centering
  \scriptsize
  \setlength{\tabcolsep}{4pt}
\begin{tabular}{lrrrr}
\toprule
\textbf{Proposer Model} & \textbf{gpt-5.4-nano:high} & \textbf{grok-4-fast-reasoning} & \textbf{qwen3.8-27b} & \textbf{Total} \\
\midrule
claude-haiku-4-5 & 680/691 (98\%) & 633/689 (92\%) & 673/689 (98\%) & 1986/2069 (96\%) \\
gpt-4o & 177/180 (98\%) & 161/180 (89\%) & 178/180 (99\%) & 516/540 (96\%) \\
grok-4-fast-reasoning & 477/506 (94\%) & 464/503 (92\%) & 453/503 (90\%) & 1394/1512 (92\%) \\
qwen/qwen3-coder & 31/36 (86\%) & 31/35 (89\%) & 34/35 (97\%) & 96/106 (91\%) \\
z-ai/glm-5.2 & 444/473 (94\%) & 351/473 (74\%) & 412/473 (87\%) & 1207/1419 (85\%) \\
qwen/qwen3.7-max & 615/659 (93\%) & 487/659 (74\%) & 563/659 (85\%) & 1665/1977 (84\%) \\
claude-sonnet-4-6 & 619/697 (89\%) & 511/697 (73\%) & 612/697 (88\%) & 1742/2091 (83\%) \\
claude-opus-4-7:low & 532/587 (91\%) & 402/587 (68\%) & 511/587 (87\%) & 1445/1761 (82\%) \\
deepseek-v3.2-thinking & 290/327 (89\%) & 226/327 (69\%) & 283/327 (87\%) & 799/981 (81\%) \\
claude-opus-4-7 & 602/679 (89\%) & 450/679 (66\%) & 591/679 (87\%) & 1643/2037 (81\%) \\
grok-4.20-0309-reasoning & 562/624 (90\%) & 413/622 (66\%) & 518/622 (83\%) & 1493/1868 (80\%) \\
gemini-3-flash-preview & 311/344 (90\%) & 176/344 (51\%) & 259/344 (75\%) & 746/1032 (72\%) \\
gpt-5.4-mini & 450/540 (83\%) & 238/540 (44\%) & 409/540 (76\%) & 1097/1620 (68\%) \\
claude-opus-4-8 & 514/688 (75\%) & 337/688 (49\%) & 526/688 (76\%) & 1377/2064 (67\%) \\
gpt-5.5:low & 209/240 (87\%) & 58/240 (24\%) & 153/240 (64\%) & 420/720 (58\%) \\
gemini-3.1-pro-preview & 425/639 (67\%) & 140/639 (22\%) & 461/639 (72\%) & 1026/1917 (54\%) \\
gpt-5.5 & 284/644 (44\%) & 135/644 (21\%) & 242/644 (38\%) & 661/1932 (34\%) \\
gpt-5.4 & 218/670 (33\%) & 13/670 (2\%) & 232/670 (35\%) & 463/2010 (23\%) \\
\midrule
\textbf{Total} & 7440/9224 (81\%) & 5226/9216 (57\%) & 7110/9216 (77\%) & 19776/27656 (72\%) \\
\bottomrule
\end{tabular}
  \label{solve-cmp}
  \end{table*}

  \begin{table*}[h]
    \caption{Puzzle outcomes by model and role. The left columns show performance in the solver role (number of puzzles solved and failed to solve). The middle columns show performance in the proposer role (number of invented puzzles solved, unsolved (\% of all correct (non-penalized) puzzles), puzzles with incorrect sample solutions which the proposer was penalized for). In a second run, we had every model try to solve its own puzzle without the prior context (i.e. without seeing its notes). ``Self S'' shows the proportion of the model's proposed puzzles that it could itself solve in the separate context, and ``Either S'' shows the proportion of puzzles that could be solved by either the model itself in a separate context or the model's opponent in the duel. This is to ensure that puzzles are solvable.}
  \centering
  \scriptsize
  \setlength{\tabcolsep}{3pt}

\begin{tabular}{l|rr|rrr|rr}
\hline
\textbf{Model} & \textbf{Solv S (\%)} & \textbf{Solv F} & \textbf{Prop S} & \textbf{Prop U (\%)} & \textbf{Prop P (\%)} & \textbf{Self S (\%)} & \textbf{Either S (\%)} \\
\hline
gpt-5.4 & 450 (98.9\%) & 5 & 239 & 431 (64.3\%) & 44 (6.2\%) & 555 (82.8\%) & 571 (85.2\%) \\
gpt-5.5 & 482 (97.6\%) & 12 & 372 & 273 (42.3\%) & 69 (9.7\%) & 618 (96.0\%) & 626 (97.2\%) \\
claude-opus-4-7 & 458 (92.0\%) & 40 & 558 & 121 (17.8\%) & 35 (4.9\%) & 621 (91.5\%) & 651 (95.9\%) \\
qwen/qwen3.7-ma... & 475 (91.5\%) & 44 & 571 & 88 (13.4\%) & 55 (7.7\%) & 614 (93.2\%) & 645 (97.9\%) \\
gpt-5.5:low & 442 (89.3\%) & 53 & 172 & 68 (28.3\%) & 474 (66.4\%) & 231 (96.2\%) & 238 (99.2\%) \\
claude-sonnet-4... & 461 (88.5\%) & 60 & 602 & 95 (13.6\%) & 17 (2.4\%) & 638 (91.5\%) & 663 (95.1\%) \\
claude-opus-4-8 & 412 (88.2\%) & 55 & 483 & 205 (29.8\%) & 26 (3.6\%) & 582 (84.6\%) & 614 (89.2\%) \\
gemini-3.1-pro-... & 424 (85.7\%) & 71 & 398 & 241 (37.7\%) & 75 (10.5\%) & 565 (88.4\%) & 593 (92.8\%) \\
claude-opus-4-7... & 437 (85.5\%) & 74 & 501 & 88 (14.9\%) & 125 (17.5\%) & 545 (92.8\%) & 568 (96.8\%) \\
gpt-5.4-mini & 392 (84.5\%) & 72 & 407 & 133 (24.6\%) & 174 (24.4\%) & 432 (80.0\%) & 485 (89.8\%) \\
z-ai/glm-5.2 & 422 (80.8\%) & 100 & 415 & 58 (12.3\%) & 241 (33.8\%) & 443 (93.7\%) & 469 (99.2\%) \\
grok-4.20-0309-... & 400 (79.8\%) & 101 & 530 & 96 (15.3\%) & 88 (12.3\%) & 554 (89.1\%) & 599 (96.3\%) \\
gemini-3-flash-... & 374 (72.9\%) & 139 & 261 & 83 (24.1\%) & 370 (51.8\%) & 297 (86.3\%) & 327 (95.1\%) \\
deepseek-v3.2-t... & 371 (71.9\%) & 145 & 278 & 49 (15.0\%) & 387 (54.2\%) & 253 (77.4\%) & 303 (92.7\%) \\
claude-haiku-4-... & 374 (69.5\%) & 164 & 678 & 13 (1.9\%) & 23 (3.2\%) & 680 (98.7\%) & 688 (99.9\%) \\
grok-4-fast-rea... & 338 (62.0\%) & 207 & 465 & 43 (8.5\%) & 206 (28.9\%) & 474 (94.2\%) & 497 (98.8\%) \\
qwen/qwen3-code... & 302 (53.4\%) & 264 & 32 & 4 (11.1\%) & 678 (95.0\%) & 32 (91.4\%) & 35 (100.0\%) \\
gpt-4o & 127 (20.8\%) & 484 & 179 & 1 (0.6\%) & 534 (74.8\%) & 162 (90.0\%) & 180 (100.0\%) \\
\hline
\textbf{Total} & 7141 (77.4\%) & 2090 & 7141 & 2090 (22.6\%) & 3621 (28.2\%) & 8296 (90.0\%) & 8752 (95.0\%) \\
\hline
\end{tabular}

  \label{combined}
  \end{table*}

\begin{table}[h]
  \centering
  \small
    \caption{Puzzle Outcomes By Turn Number}
\begin{tabular}{rrrr}
\hline
\textbf{Turn Number} & \textbf{Solved} & \textbf{Unsolved} & \textbf{Penalty (\%)} \\
\hline
1 & 594 & 117 & 207 (22.5\%) \\
2 & 540 & 133 & 245 (26.7\%) \\
3 & 533 & 139 & 246 (26.8\%) \\
4 & 517 & 135 & 266 (29.0\%) \\
5 & 507 & 146 & 265 (28.9\%) \\
6 & 480 & 160 & 278 (30.3\%) \\
7 & 510 & 148 & 260 (28.3\%) \\
8 & 492 & 167 & 259 (28.2\%) \\
9 & 511 & 150 & 257 (28.0\%) \\
10 & 478 & 162 & 278 (30.3\%) \\
11 & 504 & 153 & 261 (28.4\%) \\
12 & 491 & 160 & 267 (29.1\%) \\
13 & 494 & 146 & 278 (30.3\%) \\
14 & 490 & 174 & 254 (27.7\%) \\
\hline
\end{tabular}
  \label{byturn}
  \end{table}

\begin{table}[htbp]
  \caption{Run-to-run stability of match outcomes. Every ordered pairing (A proposes first, against B) is repeated three times in our tournament. Per model, over the pairings it took part in: \emph{Identical} is the share of pairings with the same outcome in all three repetitions; \emph{Reversed} is the share where the model both won and lost the same pairing across repetitions; \emph{Score var.}\ is the mean within-pairing variance of the match score (win $=1$, draw $=0.5$, loss $=0$) --- a fair coin between win and loss would give $0.25$. Overall, 61\% of pairings are perfectly repeatable and only 18\% ever reverse; most residual variability is between a win and a draw.}
  \label{tab:stability}
  \centering
  \footnotesize
  \begin{tabular}{lrrrr}
\toprule
\textbf{Model} & \textbf{Pairings} & \textbf{Identical (\%)} & \textbf{Reversed (\%)} & \textbf{Score var.} \\
\midrule
gpt-4o & 34 & 97.1 & 0.0 & 0.002 \\
qwen/qwen3-coder & 34 & 91.2 & 0.0 & 0.005 \\
gpt-5.4 & 34 & 88.2 & 5.9 & 0.016 \\
deepseek-v3.2-thinking & 34 & 67.6 & 17.6 & 0.044 \\
gpt-5.5 & 34 & 67.6 & 17.6 & 0.044 \\
gemini-3.1-pro-preview & 34 & 64.7 & 26.5 & 0.059 \\
gpt-5.5:low & 34 & 61.8 & 14.7 & 0.046 \\
gemini-3-flash-preview & 34 & 58.8 & 23.5 & 0.059 \\
grok-4-fast-reasoning & 34 & 55.9 & 20.6 & 0.054 \\
claude-opus-4-8 & 34 & 52.9 & 23.5 & 0.062 \\
gpt-5.4-mini & 34 & 52.9 & 32.4 & 0.080 \\
grok-4.20-0309-reasoning & 34 & 52.9 & 32.4 & 0.077 \\
qwen/qwen3.7-max & 34 & 52.9 & 8.8 & 0.041 \\
claude-haiku-4-5 & 34 & 50.0 & 23.5 & 0.062 \\
z-ai/glm-5.2 & 34 & 50.0 & 29.4 & 0.072 \\
claude-opus-4-7:low & 34 & 47.1 & 14.7 & 0.054 \\
claude-opus-4-7 & 34 & 44.1 & 11.8 & 0.049 \\
claude-sonnet-4-6 & 34 & 44.1 & 14.7 & 0.051 \\
\midrule
\textbf{Total} & 306 & 61.1 & 17.6 & 0.049 \\
\bottomrule
\end{tabular}

\end{table}

\begin{table*}[htbp]
  \caption{Robustness of the Spearman correlations between TTG Elo ratings (match, proposer, and solver Elo) and external benchmarks. For each pair we report the correlation $\rho$, a bootstrap 95\% confidence interval (2000 resamples over models), the raw $p$-value alongside Bonferroni- and Benjamini-Hochberg-adjusted values across all 15 tests, and a leave-one-out (LOO) sensitivity analysis: the range of $\rho$ when any single model is removed, the model whose removal moves $\rho$ the most (with the resulting $\rho$), and whether the correlation remains significant ($p<0.05$) in every LOO subset.}
  \label{tab:robustness}
  \centering
  \scriptsize
  \setlength{\tabcolsep}{2pt}
  \begin{tabular}{ll r r c r r r c l c}
\toprule
\textbf{Elo} & \textbf{Bench.} & $n$ & $\rho$ & \textbf{95\% CI} & $p$ & $p_{\mathrm{Bonf}}$ & $p_{\mathrm{BH}}$ & \textbf{LOO $\rho$ range} & \textbf{Most infl.\ model ($\rho$ w/o)} & \textbf{sig.\ all LOO} \\
\midrule
Match & HLE & 14 & $+0.688$ & $[+0.26, +0.91]$ & 0.0065 & 0.0981 & 0.0070 & $[+0.61, +0.78]$ & z-ai/glm-5.2 ($+0.78$) & yes \\
 & ARC-AGI & 13 & $+0.835$ & $[+0.50, +0.97]$ & 0.0004 & 0.0057 & 0.0007 & $[+0.79, +0.89]$ & gemini-3-flash... ($+0.89$) & yes \\
 & SWE-BP & 14 & $+0.719$ & $[+0.36, +0.90]$ & 0.0038 & 0.0567 & 0.0047 & $[+0.65, +0.78]$ & gpt-4o ($+0.65$) & yes \\
 & TQ & 14 & $+0.767$ & $[+0.31, +0.96]$ & 0.0014 & 0.0205 & 0.0021 & $[+0.71, +0.84]$ & gemini-3-flash... ($+0.84$) & yes \\
 & GPQA-D & 17 & $+0.771$ & $[+0.42, +0.93]$ & 0.0003 & 0.0044 & 0.0007 & $[+0.72, +0.81]$ & gpt-4o ($+0.72$) & yes \\
\addlinespace[2pt]
Proposer & HLE & 14 & $+0.688$ & $[+0.24, +0.91]$ & 0.0065 & 0.0981 & 0.0070 & $[+0.61, +0.78]$ & z-ai/glm-5.2 ($+0.78$) & yes \\
 & ARC-AGI & 13 & $+0.835$ & $[+0.52, +0.97]$ & 0.0004 & 0.0057 & 0.0007 & $[+0.79, +0.89]$ & gemini-3-flash... ($+0.89$) & yes \\
 & SWE-BP & 14 & $+0.719$ & $[+0.31, +0.91]$ & 0.0038 & 0.0567 & 0.0047 & $[+0.65, +0.78]$ & gpt-4o ($+0.65$) & yes \\
 & TQ & 14 & $+0.767$ & $[+0.33, +0.95]$ & 0.0014 & 0.0205 & 0.0021 & $[+0.71, +0.84]$ & gemini-3-flash... ($+0.84$) & yes \\
 & GPQA-D & 17 & $+0.768$ & $[+0.40, +0.93]$ & 0.0003 & 0.0048 & 0.0007 & $[+0.72, +0.81]$ & gpt-4o ($+0.72$) & yes \\
\addlinespace[2pt]
Solver & HLE & 14 & $+0.815$ & $[+0.43, +0.96]$ & 0.0004 & 0.0057 & 0.0007 & $[+0.77, +0.89]$ & z-ai/glm-5.2 ($+0.89$) & yes \\
 & ARC-AGI & 13 & $+0.901$ & $[+0.64, +1.00]$ & 0.0000 & 0.0004 & 0.0002 & $[+0.87, +0.93]$ & grok-4.20-0309... ($+0.93$) & yes \\
 & SWE-BP & 14 & $+0.684$ & $[+0.26, +0.90]$ & 0.0070 & 0.1056 & 0.0070 & $[+0.60, +0.73]$ & gpt-4o ($+0.60$) & yes \\
 & TQ & 14 & $+0.873$ & $[+0.56, +0.98]$ & 0.0000 & 0.0007 & 0.0002 & $[+0.84, +0.92]$ & grok-4.20-0309... ($+0.92$) & yes \\
 & GPQA-D & 17 & $+0.870$ & $[+0.58, +0.97]$ & 0.0000 & 0.0001 & 0.0001 & $[+0.84, +0.91]$ & gpt-5.4-mini ($+0.91$) & yes \\
\bottomrule
\end{tabular}

\end{table*}

\section{Cost}
\label{app:cost}

Table~\ref{cost} shows the cost of our tournament (all 918 duels).

\begin{table}[h]
  \centering
  \footnotesize
  \setlength{\tabcolsep}{4pt}
    \caption{Cost of running our 918 duels.}

\begin{tabular}{l|rrrrr|r}
\toprule
\textbf{Model} & \textbf{In \$/M} & \textbf{Out \$/M} & \textbf{In tokens} & \textbf{Out tokens} & \textbf{Calls} & \textbf{Cost (USD)} \\
\midrule
claude-haiku-4-5 & 1.00 & 5.00 & 2\,011\,860 & 17\,406\,364 & 1252 & \$89.04 \\
claude-opus-4-7 & 5.00 & 25.00 & 2\,343\,607 & 8\,134\,940 & 1212 & \$215.09 \\
claude-opus-4-7:low & 5.00 & 25.00 & 2\,176\,891 & 3\,328\,092 & 1225 & \$94.09 \\
claude-opus-4-8 & 5.00 & 25.00 & 2\,779\,193 & 8\,774\,103 & 1181 & \$233.25 \\
claude-sonnet-4-6 & 3.00 & 15.00 & 2\,323\,996 & 20\,482\,467 & 1235 & \$314.21 \\
deepseek-v3.2-thinking & 0.27 & 0.40 & 1\,674\,783 & 17\,535\,380 & 1230 & \$7.46 \\
gemini-3-flash-preview & 0.50 & 3.00 & 2\,432\,448 & 19\,504\,634 & 1227 & \$59.73 \\
gemini-3.1-pro-preview & 2.00 & 12.00 & 2\,671\,267 & 26\,454\,751 & 1209 & \$322.80 \\
gpt-4o & 2.50 & 10.00 & 1\,691\,330 & 463\,887 & 1325 & \$8.87 \\
gpt-5.4 & 2.50 & 15.00 & 3\,527\,768 & 26\,120\,515 & 1169 & \$400.63 \\
gpt-5.4-mini & 0.75 & 4.50 & 2\,307\,606 & 14\,188\,898 & 1178 & \$65.58 \\
gpt-5.5 & 5.00 & 30.00 & 3\,332\,889 & 10\,766\,055 & 1208 & \$339.65 \\
gpt-5.5:low & 5.00 & 30.00 & 1\,790\,731 & 1\,396\,065 & 1209 & \$50.84 \\
grok-4-fast-reasoning & 0.20 & 0.50 & 1\,476\,870 & 2\,749\,291 & 1259 & \$1.67 \\
grok-4.20-0309-reasoning & 1.25 & 2.50 & 1\,692\,488 & 11\,451\,543 & 1215 & \$30.74 \\
qwen/qwen3-coder & 0.30 & 1.00 & 2\,044\,030 & 3\,083\,135 & 1280 & \$3.70 \\
qwen/qwen3.7-max & 1.48 & 4.42 & 2\,212\,006 & 17\,707\,516 & 1233 & \$81.62 \\
z-ai/glm-5.2 & 0.65 & 2.04 & 1\,805\,295 & 13\,636\,136 & 1236 & \$29.01 \\
\midrule
\textbf{Total} & & & 40\,295\,058 & 223\,183\,772 & & \textbf{\$2347.97} \\
\bottomrule
\end{tabular}

  \vspace{-1.5em}
\label{cost}
\end{table}

\section{Sources for public results on existing benchmarks}
\label{app:benchmark-results}

For HLE, TextQuests, SWE-Bench Pro, and ARC-AGI, we used numbers from the Center for AI Safety's AI leaderboard (\url{https://dashboard.safe.ai/}). For GPQA Diamond, we used the leaderboard provided by Artificial Analysis (\url{https://artificialanalysis.ai/evaluations/gpqa-diamond}).

\FloatBarrier
\section{Puzzle Examples}
\label{app:examples}

Here we show several interesting examples of puzzles designed and solved by various models.

\noindent The examples in this appendix:
\begin{itemize}
\setlength\itemsep{0pt}
\item \ref{ex:mersenne}\ \ Mersenne Trap
\item \ref{ex:pell61}\ \ Fermat's Pell Equation
\item \ref{ex:sqtri}\ \ Square-Triangular Numbers
\item \ref{ex:minirsa}\ \ Mini-RSA Self-Fail
\item \ref{ex:rsafix}\ \ RSA Fixed Point
\item \ref{ex:hexdigit}\ \ One Hallucinated Hex Digit
\item \ref{ex:genesis}\ \ Bitcoin Genesis Nonce
\item \ref{ex:exponents}\ \ Password in the Exponents
\item \ref{ex:alphaquine}\ \ The Pure-Alphabetic Quine
\item \ref{ex:novelquine}\ \ A Novel Quine
\item \ref{ex:quine49}\ \ The 49-Character Quine
\item \ref{ex:backtick}\ \ Solved but Lost
\item \ref{ex:brainfuck}\ \ Brainfuck Compression
\item \ref{ex:bytecode}\ \ Writing Machine Code
\item \ref{ex:church}\ \ Church Numeral 65536
\item \ref{ex:negzero}\ \ The Unsatisfiable Puzzle
\item \ref{ex:hashneg}\ \ A Hash Oracle
\end{itemize}

\subsection{Mersenne Trap}
\label{ex:mersenne}

Six innocuous lines conceal an integer factorization problem; the intended divisor is the Mersenne prime $2^{127}-1$.

\begin{python}[basicstyle=\scriptsize,breaklines=true]
def mystery(x):
    if type(x) is not int or x <= 0 or x.bit_length() != 127:
        return False
    q, r = divmod(
        170481465828581158479375834498455752437849622286529887237097945306493374451,
        x,
    )
    return r == 0 and q.bit_length() == 120
\end{python}

\noindent
\textbf{Proposed by:} gpt-5.4 \quad
\textbf{Solved by:} grok-4.20-0309-reasoning \\[0.3em]
\textbf{Outcome:} \textcolor{red}{Solver Failed} \quad
\textbf{Attempt:} \texttt{20448918011133039700492870036831\allowbreak{}58358652921595552010885122138276\allowbreak{}222100969} \quad
\textbf{Solution:} \texttt{170141183460469231731687303715884105727}

\subsection{Fermat's Pell Equation}
\label{ex:pell61}

The proposer correctly recalls the theory behind Fermat's 1657 challenge, then loses the round to a big-integer multiplication error.

\begin{python}[basicstyle=\scriptsize,breaklines=true]
def mystery(x):
    if type(x) is not str:
        return False
        
    try:
        u, v = map(int, x.split(','))
    except ValueError:
        return False
        
    if u <= 0 or v <= 10**17:
        return False
        
    return u**2 - 61 * v**2 == 1
\end{python}

\noindent
\textbf{Proposed by:} gemini-3.1-pro-preview \quad
\textbf{Solved by:} gpt-5.5 \\[0.3em]
\textbf{Outcome:} \textcolor{red}{Sample Solution Incorrect} \quad
\textbf{Attempt:} \texttt{---} \quad
\textbf{Solution:} \texttt{"6239781660251608001,798920404068305040"}

\subsection{Square-Triangular Numbers}
\label{ex:sqtri}

Neither the checker nor the prompt mentions Pell equations, yet both proposer and solver independently reach for them.

\begin{python}[basicstyle=\scriptsize,breaklines=true]
def mystery(x):
    if type(x) is not int or isinstance(x, bool):
        return False

    # x must live in this large window
    if not (10**18 < x < 10**60):
        return False

    def _isqrt(n):
        if n < 0:
            return -1
        if n == 0:
            return 0
        r = 1 << ((n.bit_length() + 1) // 2)
        while True:
            nr = (r + n // r) // 2
            if nr >= r:
                break
            r = nr
        return r

    # first structural gate
    a = _isqrt(x)
    if a * a != x:
        return False

    # second structural gate
    t = 8 * x + 1
    b = _isqrt(t)
    if b * b != t:
        return False

    return True
\end{python}

\noindent
\textbf{Proposed by:} claude-opus-4-8 \quad
\textbf{Solved by:} gpt-5.4 \\[0.3em]
\textbf{Outcome:} \textcolor{green!60!black}{Solved} \quad
\textbf{Attempt:} \texttt{2507180834294496361} \quad
\textbf{Solution:} \texttt{\_\_import\_\_(\allowbreak{}'functools')\allowbreak{}.reduce(\allowbreak{}lambda t,\allowbreak{}\_:\allowbreak{}(\allowbreak{}t[\allowbreak{}1]\allowbreak{},\allowbreak{}34*\allowbreak{}t[\allowbreak{}1]\allowbreak{}-t[\allowbreak{}0]\allowbreak{}+\allowbreak{}2)\allowbreak{},\allowbreak{}range(\allowbreak{}20)\allowbreak{},\allowbreak{}(\allowbreak{}0,\allowbreak{}1)\allowbreak{})\allowbreak{}[\allowbreak{}0]\allowbreak{}}

\subsection{Mini-RSA Self-Fail}
\label{ex:minirsa}

A five-line RSA puzzle is invalidated because the proposer cannot do its own modular exponentiation in its head.

\begin{python}[basicstyle=\scriptsize,breaklines=true]
def mystery(x):
    if isinstance(x, str):
        try:
            x = int(x)
        except ValueError:
            return False
    if not isinstance(x, int):
        return False
    n = 1022117
    e = 5
    c = 754241
    return pow(x, e, n) == c
\end{python}

\noindent
\textbf{Proposed by:} deepseek-v3.2-thinking \quad
\textbf{Solved by:} claude-haiku-4-5 \\[0.3em]
\textbf{Outcome:} \textcolor{red}{Sample Solution Incorrect} \quad
\textbf{Attempt:} \texttt{---} \quad
\textbf{Solution:} \texttt{123456}

\subsection{RSA Fixed Point}
\label{ex:rsafix}

The intended solution requires CRT reasoning about fixed points; the solver instead notices that $x=1$ satisfies the spec.

\begin{python}[basicstyle=\scriptsize,breaklines=true]
def mystery(x):
    # Convert to integer if possible
    if isinstance(x, str):
        try:
            x = int(x)
        except ValueError:
            return False
    elif not isinstance(x, int):
        return False
    if x <= 0:
        return False
    n = 1000036000099
    e = 65537
    # Check if x^e mod n equals x
    return pow(x, e, n) == x
\end{python}

\noindent
\textbf{Proposed by:} deepseek-v3.2-thinking \quad
\textbf{Solved by:} gpt-5.5 \\[0.3em]
\textbf{Outcome:} \textcolor{green!60!black}{Solved} \quad
\textbf{Attempt:} \texttt{1} \quad
\textbf{Solution:} \texttt{233341700023}

\subsection{One Hallucinated Hex Digit}
\label{ex:hexdigit}

The construction is mathematically sound, but the proposer reproduces a 64-digit curve constant from memory with a single wrong digit.

\begin{python}[basicstyle=\scriptsize,breaklines=true]
def mystery(x):
    if type(x) is not int:
        return False
    if not (0 < x < 2**256):
        return False

    m = 0xFFFFFFFFFFFFFFFFFFFFFFFFFFFFFFFFFFFFFFFFFFFFFFFFFFFFFFFEFFFFFC2F

    P = (
        0xC6047F9441ED7D6D3045406E95C07CD85AEEF0F32823C82850257530949F2818,
        0x1AE168FEA63DC339A3C58419466CEAEEF7F632653266D0E1236431A950CFE52A,
    )
    Q = (
        0xF9308A019258C31049344F85F89D5229B531C845836F99B08601F113BCE036F9,
        0x388F7B0F632DE8140FE337E62A37F3566500A99934C2231B6CB9FD7584B8E672,
    )

    def add(A, B):
        if A is None:
            return B
        if B is None:
            return A

        ax, ay = A
        bx, by = B

        if ax == bx and (ay + by) % m == 0:
            return None

        if A == B:
            s = (3 * ax * ax * pow(2 * ay, -1, m)) % m
        else:
            s = ((by - ay) * pow(bx - ax, -1, m)) % m

        cx = (s * s - ax - bx) % m
        cy = (s * (ax - cx) - ay) % m
        return (cx, cy)

    def mul(k, A):
        R = None
        while k:
            if k & 1:
                R = add(R, A)
            A = add(A, A)
            k >>= 1
        return R

    return mul(x, P) == Q
\end{python}

\noindent
\textbf{Proposed by:} gpt-5.5 \quad
\textbf{Solved by:} claude-haiku-4-5 \\[0.3em]
\textbf{Outcome:} \textcolor{red}{Sample Solution Incorrect} \quad
\textbf{Attempt:} \texttt{---} \quad
\textbf{Solution:} \texttt{0x7FFFFFFFFFFFFFFFFFFFFFFFFFFFFF\allowbreak{}FF5D576E7357A4501DDFE92F46681B20\allowbreak{}A2}

\subsection{Bitcoin Genesis Nonce}
\label{ex:genesis}

The puzzle is the genesis block header in disguise; the solver recalls Satoshi's nonce rather than brute-forcing $2^{32}$ hashes. Memorization-based puzzles do occur, but this round shows why they are usually a poor strategy: whatever constants one model has memorized, another model may have memorized too.

\begin{python}[basicstyle=\scriptsize,breaklines=true]
import hashlib

def mystery(x):
    if not isinstance(x, int) or isinstance(x, bool):
        return False
    if not (0 <= x < 2**32):
        return False

    prefix = bytes.fromhex(
        "01000000"
        "0000000000000000000000000000000000000000000000000000000000000000"
        "3ba3edfd7a7b12b27ac72c3e67768f617fc81bc3888a51323a9fb8aa4b1e5e4a"
        "29ab5f49"
        "ffff001d"
    )

    h = hashlib.sha256(hashlib.sha256(prefix + x.to_bytes(4, "little")).digest()).digest()
    return h[::-1].hex() == (
        "000000000019d668"
        "9c085ae165831e934"
        "ff763ae46a2a6c17"
        "2b3f1b60a8ce26f"
    )
\end{python}

\noindent
\textbf{Proposed by:} gpt-5.5 \quad
\textbf{Solved by:} claude-opus-4-7 \\[0.3em]
\textbf{Outcome:} \textcolor{green!60!black}{Solved} \quad
\textbf{Attempt:} \texttt{2083236893} \quad
\textbf{Solution:} \texttt{2083236893}

\subsection{Password in the Exponents}
\label{ex:exponents}

The solver peels three layers---factorization encoding, ROT13, and a known RFC test vector---to reconstruct a 200-digit integer.

\begin{python}[basicstyle=\scriptsize,breaklines=true]
def mystery(x):
    import codecs
    import hashlib
    import hmac

    if type(x) is not int:
        return False

    if x <= 1:
        return False

    primes = (
        2, 3, 5, 7, 11, 13, 17, 19,
        23, 29, 31, 37, 41, 43, 47, 53,
        59, 61, 67, 71, 73, 79, 83, 89,
    )

    raw = bytearray()
    for p in primes:
        e = 0
        while x % p == 0:
            x //= p
            e += 1
            if e > 122:
                return False

        if not (65 <= e <= 90 or 97 <= e <= 122):
            return False
        raw.append(e)

    if x != 1:
        return False

    if (sum(raw), raw[0] ^ raw[-1], raw.count(ord("f"))) != (2237, 0x12, 4):
        return False

    s = raw.decode("ascii")
    y = codecs.decode(s, "rot_13").encode("ascii")

    if y[:8] != y[16:] or y[:8].upper() != y[8:16]:
        return False

    want = bytes.fromhex(
        "3d2eec4fe41c849b80c8d83662c0e44a"
        "8b291a964cf2f07038"
    )

    got = hashlib.pbkdf2_hmac(
        "sha1",
        y,
        b"saltSALTsaltSALTsaltSALTsaltSALTsalt",
        4096,
        len(want),
    )

    return hmac.compare_digest(got, want)
\end{python}

\noindent
\textbf{Proposed by:} gpt-5.5 \quad
\textbf{Solved by:} claude-sonnet-4-6 \\[0.3em]
\textbf{Outcome:} \textcolor{green!60!black}{Solved} \quad
\textbf{Attempt:} \texttt{2*\allowbreak{}*\allowbreak{}99 *\allowbreak{} 3*\allowbreak{}*\allowbreak{}110 *\allowbreak{} 5*\allowbreak{}*\allowbreak{}102 *\allowbreak{} 7*\allowbreak{}*\allowbreak{}102 *\allowbreak{} 11*\allowbreak{}*\allowbreak{}106 *\allowbreak{} 13*\allowbreak{}*\allowbreak{}98 *\allowbreak{} 17*\allowbreak{}*\allowbreak{}101 *\allowbreak{} 19*\allowbreak{}*\allowbreak{}113 *\allowbreak{} 23*\allowbreak{}*\allowbreak{}67 *\allowbreak{} 29*\allowbreak{}*\allowbreak{}78 *\allowbreak{} 31*\allowbreak{}*\allowbreak{}70 *\allowbreak{} 37*\allowbreak{}*\allowbreak{}70 *\allowbreak{} 41*\allowbreak{}*\allowbreak{}74 *\allowbreak{} 43*\allowbreak{}*\allowbreak{}66 *\allowbreak{} 47*\allowbreak{}*\allowbreak{}69 *\allowbreak{} 53*\allowbreak{}*\allowbreak{}81 *\allowbreak{} 59*\allowbreak{}*\allowbreak{}99 *\allowbreak{} 61*\allowbreak{}*\allowbreak{}110 *\allowbreak{} 67*\allowbreak{}*\allowbreak{}102 *\allowbreak{} 71*\allowbreak{}*\allowbreak{}102 *\allowbreak{} 73*\allowbreak{}*\allowbreak{}106 *\allowbreak{} 79*\allowbreak{}*\allowbreak{}98 *\allowbreak{} 83*\allowbreak{}*\allowbreak{}101 *\allowbreak{} 89*\allowbreak{}*\allowbreak{}113} \quad
\textbf{Solution:} \texttt{2*\allowbreak{}*\allowbreak{}99*\allowbreak{}3*\allowbreak{}*\allowbreak{}110*\allowbreak{}5*\allowbreak{}*\allowbreak{}102*\allowbreak{}7*\allowbreak{}*\allowbreak{}102*\allowbreak{}11*\allowbreak{}*\allowbreak{}106*\allowbreak{}13*\allowbreak{}*\allowbreak{}98*\allowbreak{}17*\allowbreak{}*\allowbreak{}101*\allowbreak{}19*\allowbreak{}*\allowbreak{}113*\allowbreak{}23*\allowbreak{}*\allowbreak{}67*\allowbreak{}29*\allowbreak{}*\allowbreak{}78*\allowbreak{}31*\allowbreak{}*\allowbreak{}70*\allowbreak{}37*\allowbreak{}*\allowbreak{}70*\allowbreak{}41*\allowbreak{}*\allowbreak{}74*\allowbreak{}43*\allowbreak{}*\allowbreak{}66*\allowbreak{}47*\allowbreak{}*\allowbreak{}69*\allowbreak{}53*\allowbreak{}*\allowbreak{}81*\allowbreak{}59*\allowbreak{}*\allowbreak{}99*\allowbreak{}61*\allowbreak{}*\allowbreak{}110*\allowbreak{}67*\allowbreak{}*\allowbreak{}102*\allowbreak{}71*\allowbreak{}*\allowbreak{}102*\allowbreak{}73*\allowbreak{}*\allowbreak{}106*\allowbreak{}79*\allowbreak{}*\allowbreak{}98*\allowbreak{}83*\allowbreak{}*\allowbreak{}101*\allowbreak{}89*\allowbreak{}*\allowbreak{}113}

\subsection{The Pure-Alphabetic Quine}
\label{ex:alphaquine}

A quine using only letters looks impossible until one notices that \texttt{eval} runs where the input variable is bound: the answer is \texttt{"x"}.

\begin{python}[basicstyle=\scriptsize,breaklines=true]
def mystery(x):
    if type(x) is not str:
        return False
        
    # Strictly alphabetic characters only. 
    # No quotes, no spaces, no digits, no underscores, or symbols!
    if not x.isalpha():
        return False
        
    try:
        # A pure-alpha quine?
        return eval(x) == x
    except Exception:
        return False
\end{python}

\noindent
\textbf{Proposed by:} gemini-3.1-pro-preview \quad
\textbf{Solved by:} gpt-4o \\[0.3em]
\textbf{Outcome:} \textcolor{red}{Solver Failed} \quad
\textbf{Attempt:} \texttt{"validnamehere" (\allowbreak{}hypothetical under permissible aligned basis discordantly discretionarily applied against formal vanilla statement boundary observations)\allowbreak{}.} \quad
\textbf{Solution:} \texttt{"x"}

\subsection{A Novel Quine}
\label{ex:novelquine}

Rather than rediscovering the intended solution, the solver invents a structurally different quine using the walrus operator.

\begin{python}[basicstyle=\scriptsize,breaklines=true]
def mystery(x):
    if not isinstance(x, str) or len(x) > 100:
        return False
        
    # The puzzle requires you to include the word "puzzle"
    if "puzzle" not in x:
        return False
        
    # I am severely restricting the tools you can use.
    # Standard quine tricks or bypassing via builtins are banned.
    bad = ["lambda", "eval", "exec", "import", "class", "def", "for", "while", 
           "join", "replace", "format", "chr", "dict", "list", "tuple", "getattr", "setattr"]
    if any(b in x for b in bad):
        return False
        
    try:
        # A sandbox with no builtins.
        # You must produce an expression that evaluates to its own source code!
        env = {"__builtins__": {}}
        return eval(x, env, env) == x
    except Exception:
        return False
\end{python}

\noindent
\textbf{Proposed by:} gemini-3.1-pro-preview \quad
\textbf{Solved by:} z-ai/glm-5.2 \\[0.3em]
\textbf{Outcome:} \textcolor{green!60!black}{Solved} \quad
\textbf{Attempt:} \texttt{"(\allowbreak{}1 or \textbackslash\{\}"puzzle\textbackslash\{\}")\allowbreak{} and (\allowbreak{}s:\allowbreak{}=\allowbreak{}'(\allowbreak{}1 or \textbackslash\{\}"puzzle\textbackslash\{\}")\allowbreak{} and (\allowbreak{}s:\allowbreak{}=\allowbreak{}\%r)\allowbreak{}\%\%s')\allowbreak{}\%s"} \quad
\textbf{Solution:} \texttt{"(\allowbreak{}puzzle:\allowbreak{}=\allowbreak{}'(\allowbreak{}puzzle:\allowbreak{}=\allowbreak{}\%r)\allowbreak{} \%\% puzzle')\allowbreak{} \% puzzle"}

\subsection{The 49-Character Quine}
\label{ex:quine49}

The proposer's own sample falls to the classic \texttt{repr} re-escaping subtlety it built the puzzle around.

\begin{python}[basicstyle=\scriptsize,breaklines=true]
def mystery(x):
    if type(x) is not str:
        return False

    if len(x) != 49:
        return False

    if not x.isascii():
        return False

    if x.count("lambda") != 2:
        return False

    if x.count("repr") != 2:
        return False

    if x.count("%") != 4:
        return False

    if "__" in x or "." in x or "import" in x:
        return False

    try:
        y = eval(x, {"__builtins__": {}, "repr": repr}, {})
    except Exception:
        return False

    return type(y) is str and y == x
\end{python}

\noindent
\textbf{Proposed by:} gpt-5.5 \quad
\textbf{Solved by:} claude-haiku-4-5 \\[0.3em]
\textbf{Outcome:} \textcolor{red}{Sample Solution Incorrect} \quad
\textbf{Attempt:} \texttt{---} \quad
\textbf{Solution:} \texttt{"(\allowbreak{}lambda s:\allowbreak{}s\%repr(\allowbreak{}s)\allowbreak{})\allowbreak{}(\allowbreak{}'(\allowbreak{}lambda s:\allowbreak{}s\%\%repr(\allowbreak{}s)\allowbreak{})\allowbreak{}(\allowbreak{}\%r)\allowbreak{}')\allowbreak{}"}

\subsection{Solved but Lost}
\label{ex:backtick}

The solver derives exactly the intended quine, wraps it in markdown backticks, and loses the round to formatting.

\begin{python}[basicstyle=\scriptsize,breaklines=true]
def mystery(x):
    if type(x) is not str or len(x) > 50:
        return False
        
    # Brutal character restriction
    allowed = set("[]s:='%r,1")
    if not set(x).issubset(allowed):
        return False
        
    try:
        # Evaluate in a strictly sandboxed environment
        res = eval(x, {'__builtins__': {}})
        
        # Enforce exact type
        if type(res) is not str:
            return False
            
        return res == x
    except Exception:
        return False
\end{python}

\noindent
\textbf{Proposed by:} gemini-3.1-pro-preview \quad
\textbf{Solved by:} claude-opus-4-8 \\[0.3em]
\textbf{Outcome:} \textcolor{red}{Solver Failed} \quad
\textbf{Attempt:} \texttt{`[s:='[s:=\%r,s\%\%s][1]',s\%s][1]`} \quad
\textbf{Solution:} \texttt{"[s:='[s:=\%r,s\%\%s][1]',s\%s][1]"}

\subsection{Brainfuck Compression}
\label{ex:brainfuck}

Printing 1000 Z's in at most 50 characters of Brainfuck forces the solver to factor the problem into nested loops.

\begin{python}[basicstyle=\scriptsize,breaklines=true]
def mystery(x):
    if not isinstance(x, str) or len(x) > 50:
        return False
        
    # Only standard Brainfuck characters allowed
    if not set(x).issubset(set("><+-.,[]")):
        return False
    
    memory = [0] * 30000
    ptr = 0
    output = []
    
    # Precompute jumps to make simulation efficient
    loop_map = {}
    stack = []
    for i, c in enumerate(x):
        if c == '[': 
            stack.append(i)
        elif c == ']':
            if not stack: return False
            start = stack.pop()
            loop_map[start] = i
            loop_map[i] = start
    if stack: 
        return False
    
    i = 0
    steps = 0
    while i < len(x):
        steps += 1
        # Prevent infinite loops
        if steps > 100000: 
            return False
            
        c = x[i]
        if c == '>': ptr = (ptr + 1) % 30000
        elif c == '<': ptr = (ptr - 1) % 30000
        elif c == '+': memory[ptr] = (memory[ptr] + 1) % 256
        elif c == '-': memory[ptr] = (memory[ptr] - 1) % 256
        elif c == '.': output.append(chr(memory[ptr]))
        elif c == '[':
            if memory[ptr] == 0: i = loop_map[i]
        elif c == ']':
            if memory[ptr] != 0: i = loop_map[i]
            
        i += 1
        
    # The program must output exactly 1000 'Z' characters
    return "".join(output) == "Z" * 1000
\end{python}

\noindent
\textbf{Proposed by:} gemini-3.1-pro-preview \quad
\textbf{Solved by:} claude-opus-4-7 \\[0.3em]
\textbf{Outcome:} \textcolor{green!60!black}{Solved} \quad
\textbf{Attempt:} \texttt{"+\allowbreak{}+\allowbreak{}+\allowbreak{}+\allowbreak{}+\allowbreak{}+\allowbreak{}+\allowbreak{}+\allowbreak{}+\allowbreak{}+\allowbreak{}[\allowbreak{}>+\allowbreak{}+\allowbreak{}+\allowbreak{}+\allowbreak{}+\allowbreak{}+\allowbreak{}+\allowbreak{}+\allowbreak{}+\allowbreak{}<-]\allowbreak{}>>+\allowbreak{}+\allowbreak{}+\allowbreak{}+\allowbreak{}[\allowbreak{}>------[\allowbreak{}<<.>>-]\allowbreak{}<-]\allowbreak{}"} \quad
\textbf{Solution:} \texttt{"+\allowbreak{}+\allowbreak{}+\allowbreak{}+\allowbreak{}+\allowbreak{}+\allowbreak{}+\allowbreak{}+\allowbreak{}+\allowbreak{}[\allowbreak{}>+\allowbreak{}+\allowbreak{}+\allowbreak{}+\allowbreak{}+\allowbreak{}+\allowbreak{}+\allowbreak{}+\allowbreak{}+\allowbreak{}+\allowbreak{}<-]\allowbreak{}>>+\allowbreak{}+\allowbreak{}+\allowbreak{}+\allowbreak{}[\allowbreak{}>------[\allowbreak{}<<.>>-]\allowbreak{}<-]\allowbreak{}"}

\subsection{Writing Machine Code}
\label{ex:bytecode}

The solver must hand-assemble 16 bytes of bytecode for the proposer's custom stack machine.

\begin{python}[basicstyle=\scriptsize,breaklines=true]
def mystery(x):
    if not isinstance(x, str) or len(x) != 32:
        return False
    try:
        b = bytes.fromhex(x)
    except ValueError:
        return False
        
    if len(b) != 16:
        return False

    stack = []
    ip = 0
    steps = 0
    
    while ip < len(b):
        steps += 1
        if steps > 200:
            return False
            
        op = b[ip]
        ip += 1
        
        if op == 0x1A: # PUSH
            if ip >= len(b): return False
            stack.append(b[ip])
            ip += 1
        elif op == 0x4B: # ADD
            if len(stack) < 2: return False
            a, b_ = stack.pop(), stack.pop()
            stack.append((a + b_) & 0xFF)
        elif op == 0x2D: # SUB
            if len(stack) < 2: return False
            a, b_ = stack.pop(), stack.pop()
            stack.append((b_ - a) & 0xFF)
        elif op == 0x8C: # MUL
            if len(stack) < 2: return False
            a, b_ = stack.pop(), stack.pop()
            stack.append((a * b_) & 0xFF)
        elif op == 0x5E: # DUP
            if len(stack) < 1: return False
            stack.append(stack[-1])
        elif op == 0xFF: # HALT
            return len(stack) == 1 and stack[0] == 0x59
        else:
            return False
    return False
\end{python}

\noindent
\textbf{Proposed by:} z-ai/glm-5.2 \quad
\textbf{Solved by:} qwen/qwen3-coder \\[0.3em]
\textbf{Outcome:} \textcolor{red}{Solver Failed} \quad
\textbf{Attempt:} \texttt{"1a005e2d1a005e4b1a595e2d1a592dff"} \quad
\textbf{Solution:} \texttt{"1a5a1a012dff00000000000000000000"}

\subsection{Church Numeral 65536}
\label{ex:church}

Restricted to the characters \texttt{lambda :()xyz}, the solver finds the same self-applying doubling tower as the proposer: $2^{2^{2^{2}}}$.

\begin{python}[basicstyle=\scriptsize,breaklines=true]
def mystery(x: str) -> bool:
    if not isinstance(x, str) or len(x) > 48:
        return False
        
    # Strictly limited to pure lambda calculus
    allowed = set("lambda :()xyz")
    if set(x) - allowed:
        return False
        
    try:
        # Evaluate x in a completely pristine namespace
        f = eval(x, {})
        
        # The evaluated expression must be a valid Church numeral
        # that computes exactly 65,536 when applied to an incrementer
        return f(lambda n: n + 1)(0) == 65536
    except Exception:
        return False
\end{python}

\noindent
\textbf{Proposed by:} gemini-3.1-pro-preview \quad
\textbf{Solved by:} gpt-5.4 \\[0.3em]
\textbf{Outcome:} \textcolor{green!60!black}{Solved} \quad
\textbf{Attempt:} \texttt{"(\allowbreak{}lambda x:\allowbreak{}x(\allowbreak{}x)\allowbreak{}(\allowbreak{}x)\allowbreak{}(\allowbreak{}x)\allowbreak{})\allowbreak{}(\allowbreak{}lambda y:\allowbreak{}lambda z:\allowbreak{}y(\allowbreak{}y(\allowbreak{}z)\allowbreak{})\allowbreak{})\allowbreak{}"} \quad
\textbf{Solution:} \texttt{"(\allowbreak{}lambda x:\allowbreak{}x(\allowbreak{}x)\allowbreak{}(\allowbreak{}x)\allowbreak{}(\allowbreak{}x)\allowbreak{})\allowbreak{}(\allowbreak{}lambda y:\allowbreak{}lambda z:\allowbreak{}y(\allowbreak{}y(\allowbreak{}z)\allowbreak{})\allowbreak{})\allowbreak{}"}

\subsection{The Unsatisfiable Puzzle}
\label{ex:negzero}

Under IEEE-754 semantics $-0.0$ is a valid answer, but Python raises on division by zero, so no input can ever succeed; the puzzle is therefore scored as invalid and the proposer is penalized.

\begin{python}[basicstyle=\scriptsize,breaklines=true]
def mystery(x):
    try:
        return x == 0 and 1 / x < 0
    except:
        return False
\end{python}

\noindent
\textbf{Proposed by:} grok-4.20-0309-reasoning \quad
\textbf{Solved by:} deepseek-v3.2-thinking \\[0.3em]
\textbf{Outcome:} \textcolor{red}{Sample Solution Incorrect} \quad
\textbf{Attempt:} \texttt{---} \quad
\textbf{Solution:} \texttt{-0.0}

\subsection{A Hash Oracle}
\label{ex:hashneg}

The only negative integer with $\mathrm{hash}(x)=2x$ is $-1$, an artifact of CPython reserving $-1$ as an error sentinel.

\begin{python}[basicstyle=\scriptsize,breaklines=true]
def mystery(x):
    if type(x) is not int or x >= 0:
        return False
    try:
        # In Python's internal logic, hashes are usually predictable.
        # However, a specific architectural constraint creates a 
        # "collision" where one integer's identity is shifted by the 
        # interpreter to avoid a reserved error state.
        return hash(x) == (x << 1)
    except Exception:
        return False
\end{python}

\noindent
\textbf{Proposed by:} gemini-3-flash-preview \quad
\textbf{Solved by:} gemini-3.1-pro-preview \\[0.3em]
\textbf{Outcome:} \textcolor{green!60!black}{Solved} \quad
\textbf{Attempt:} \texttt{-1} \quad
\textbf{Solution:} \texttt{-1}

\section{Prompts}

The following is the prompt template we show the proposer model:

\begingroup\small
\begin{verbatim}
Let's play a game.

# Rules
We take turns.
One player, the proposer, gives the other a code puzzle with a
Python function called `mystery` that takes a value x (either
string or number) and returns a boolean.
The other player, the solver, have to find a value x for
which the function returns True.
The proposer's job is to make the puzzle as hard as they can,
i.e. it should be very difficult to find out which
values the function returns True for.
To prove that the puzzle is solvable, the proposer must give a
sample solution.
If the sample solution turns out to be wrong, the proposer is
penalized.

# Instructions
You are the proposer. I am the solver.

Respond with a Python code block first, then explain your puzzle.
Everything outside the code block is private -- I cannot see it.
The last line of your response should contain only: `SOLUTION: x`
where x is the value of a sample solution for your puzzle.
Ensure that `mystery(x)` is valid Python code given your x.
For example, it is very important to use quotes if you believe
`x` is a string.

Please make sure that your sample solution works -- you will be
penalized if not.
Do not give multiple code blocks or puzzles.

# History
Here are the puzzles from past turns of the game.
You should NOT repeat them,
but you can learn from them to ensure your puzzle is difficult
enough.

{past}

# Reminder
Please ensure your response ends with this line: `SOLUTION: x`
where x is the value of a sample solution for your puzzle.
\end{verbatim}
\endgroup

The following is the much shorter prompt for the solver:

\begingroup\small
\begin{verbatim}
Here's a Python function that takes a value x and returns a boolean.
Please give me a value for x such that mystery(x) is True.
The last line of your response should be: `SOLUTION: x`
where x is the value. Ensure that `mystery(x)` is valid Python code given your x.
For example, if you believe `x` is a string, use quotes.
Example: `SOLUTION: \"Hello, world!\"`

```python
{puzzle}
```
\end{verbatim}
\endgroup



\end{document}